\documentclass[11pt]{article}

\usepackage[margin=1in]{geometry}
\usepackage{amsmath,amssymb,amsfonts,mathtools}
\usepackage{graphicx}
\usepackage{booktabs}
\usepackage{multirow}
\usepackage{array}
\usepackage{enumitem}
\usepackage{xcolor}
\usepackage{url}
\usepackage{microtype}
\usepackage{natbib}
\usepackage[hidelinks]{hyperref}
\usepackage[nameinlink,capitalize]{cleveref}
\usepackage{placeins}

\usepackage{algorithm}
\usepackage{algpseudocode}
\usepackage{amsthm}

\usepackage{xcolor}

\theoremstyle{definition}

\title{CLEAR: Cross-Source Evidence Adjudication for Large Language Models in Medicine}

\author{
Shuai Wang$^{1}$ \quad
Yize Zhao$^{1,*}$ \quad
Qingyu Chen$^{2,*}$\\[0.6em]
\parbox{0.92\textwidth}{
\centering\small
$^{1}$Department of Biostatistics, Yale School of Public Health,
Yale University, New Haven, CT, USA\\
$^{2}$Department of Biomedical Informatics and Data Science,
Yale School of Medicine, Yale University, New Haven, CT, USA
}
}

\date{}

\date{}

\begin{document}

\maketitle

\begingroup
\renewcommand{\thefootnote}{*}
\footnotetext{Co-corresponding authors:
\texttt{yize.zhao@yale.edu} and
\texttt{qingyu.chen@yale.edu}}
\endgroup

\begin{abstract}
Medical knowledge evolves continuously, whereas the parametric knowledge encoded in large language models (LLMs) is fixed at training time. External retrieval, including retrieval-augmented generation (RAG), can provide access to newly available evidence, but retrieved information may be irrelevant, incomplete, or conflicting. As a result, external retrieval can in turn degrade the factual accuracy and evidence grounding of LLM outputs. To address this challenge, we propose \textbf{CLEAR}, an agentic framework for cross-source evidence adjudication in LLMs in medicine. CLEAR independently generates candidate answers from three complementary pathways---parametric knowledge, locally curated corpora, and dynamically retrieved evidence---reflecting three common sources of information available to LLMs. An aggregation verifier jointly evaluates the candidates, supporting evidence, provenance, and source-quality information to identify agreement and conflict across sources. An adjudication module then determines whether the current conclusion should be preserved or revised through complementary override-guard and challenge-audit mechanisms, while unresolved conflicts trigger targeted follow-up search and re-adjudication.

We evaluate CLEAR on ten public benchmarks spanning multiple-choice and free-text settings using representative commercial and open-weight backbones (o3-mini and Qwen-3.5-9B). We compare CLEAR with direct inference and established RAG baselines under matched backbone settings. CLEAR achieves competitive performance across the ten benchmarks, with the largest gains observed in settings where direct inference or conventional retrieval provides a weaker baseline. In settings where parametric knowledge already supports strong performance, CLEAR generally remains competitive, whereas additional retrieval does not consistently improve performance. With Qwen-3.5-9B, the best CLEAR configurations improve performance by 11.30 percentage points on NEJM-QA and 13.69 percentage points on MedRBench over the strongest evaluated baselines. Ablation studies further show that the full cross-source framework achieves stronger aggregate performance than individual evidence-source settings, while targeted follow-up search provides additional gains on unresolved cases. These findings suggest that an agentic framework can benefit LLMs not only by acquiring new evidence, but also by explicitly adjudicating conflicts across heterogeneous knowledge sources.
\end{abstract}

\section{Introduction}
\label{sec:introduction}

Large language models (LLMs) have shown remarkable potential across a wide range of knowledge-intensive tasks, demonstrating strong capabilities in language understanding, question answering, and complex reasoning~\cite{brown2020language,chowdhery2023palm,llama2023llama,achiam2023gpt}. By acquiring broad world knowledge from large-scale corpora during pretraining, these models provide a general foundation that can be further adapted to downstream domains. These capabilities have extended to medicine, where medically adapted LLMs have shown promise in medical knowledge and reasoning, disease diagnosis, and patient communication~\cite{tian2024opportunities,liu2025application,chen2025benchmarking,chen2026llm}. Taking medical question answering as an example, Med-PaLM achieved 67.6\% accuracy on MedQA and increased the proportion of long-form answers judged to be aligned with scientific consensus from 61.9\% for Flan-PaLM to 92.6\%~\cite{singhal2023large}. Med-PaLM 2 further increased MedQA accuracy to 86.5\%, approaching expert-level performance on medical licensing examination questions~\cite{singhal2025toward}. Open-weight medical LLMs have likewise demonstrated competitive performance across multiple medical benchmarks, indicating that strong medical reasoning capabilities are not limited to proprietary models~\cite{chen2024meditron,bolton2024biomedlm,xie2025medical}.

However, the medical knowledge encoded in LLM parameters is inherently bounded by the data available during training. After training is completed, a model does not automatically incorporate evidence published thereafter~\cite{dhingra2022time}. Medical evidence, by contrast, evolves rapidly through newly published clinical trials, updated practice guidelines, new drug approvals, and emerging safety signals. In 2024, 289 clinical practice guidelines were published, representing a 29\% increase from 2023~\cite{guidelinecentral2025review}. PubMed now contains more than 40 million biomedical citations, with more than one million new records added annually~\cite{nlm2024medline,kim2026medpmc}. Consequently, LLMs may generate answers that are consistent with the evidence available at the time of training but are no longer aligned with the latest clinical evidence or recommendations. This mismatch may be particularly consequential in high-stakes medical applications, for example, when a model generates treatment recommendations that conflict with updated clinical guidelines~\cite{artsi2025challenges,guan2026large,wang2026reasoning}.

To address this challenge, previous studies have commonly adopted retrieval-augmented generation (RAG), which supplements the knowledge encoded in model parameters with external evidence retrieved at inference time~\cite{amugongo2025retrieval,liu2025improving,yang2026retrieval}. By retrieving relevant information from biomedical literature, medical textbooks, clinical guidelines, and other curated knowledge resources, RAG enables LLMs to access specialized or newly published medical evidence without repeatedly retraining the underlying model~\cite{zakka2024almanac,xiong2024benchmarking}. Recent studies, however, have identified important limitations of RAG in medical applications~\cite{kim2025rethinking,xie2025retrieval,wong2025retrieval}. Retrieval may return irrelevant, low-quality, or question-misaligned passages. Moreover, LLMs may fail to identify or correctly use relevant evidence even when it has been successfully retrieved. These failures can weaken evidence attribution, reduce factuality, and degrade the accuracy of the final output~\cite{kim2025rethinking}.

These limitations point to a broader challenge: in medicine, LLMs must reason over multiple knowledge and evidence sources that differ in coverage, timeliness, authority, and reliability. We broadly categorize these sources into three types. First, \textit{parametric knowledge} refers to information acquired by an LLM during pretraining or post-training. It provides broad medical priors but may be incomplete, outdated, or insufficiently specific to the question at hand. Second, \textit{locally curated evidence}, such as institutional documents or medical textbooks, may be authoritative and task-specific but is constrained by the coverage and time of corpus construction. Third, \textit{dynamic evidence}, such as information obtained through query-time web search, may capture recent or context-specific developments but can vary substantially in relevance, credibility, and quality.

These sources may provide complementary support, but they may also lead to conflicting conclusions. Prior studies have shown that effectively coordinating internal knowledge and retrieved evidence remains an open challenge~\cite{xiong2024benchmarking,tan2025dynamic,guan2026large}. For example, a recent analysis found that conflicts between internal and external knowledge occurred in 19.2\% of evaluated cases. Among these conflicts, internal knowledge was correct in 47.4\% of cases, whereas external evidence was correct in 52.6\%~\cite{wang2025astute}. These findings suggest that neither source can be trusted by default. Recent extensions of RAG have introduced adaptive retrieval, query reformulation, and follow-up question generation, allowing models to determine whether additional retrieval is needed and how the search process should proceed~\cite{sohn2025rationale,xiong2024improving}. However, these methods primarily focus on whether and how to retrieve additional evidence rather than on resolving conflicts among heterogeneous sources after retrieval. A central challenge therefore remains: how to adjudicate heterogeneous sources and determine whether the system should preserve an existing conclusion, revise it in light of newly acquired evidence, or seek additional evidence when the conflict remains unresolved.

\begin{figure}[t]
    \centering
    \includegraphics[width=\textwidth]{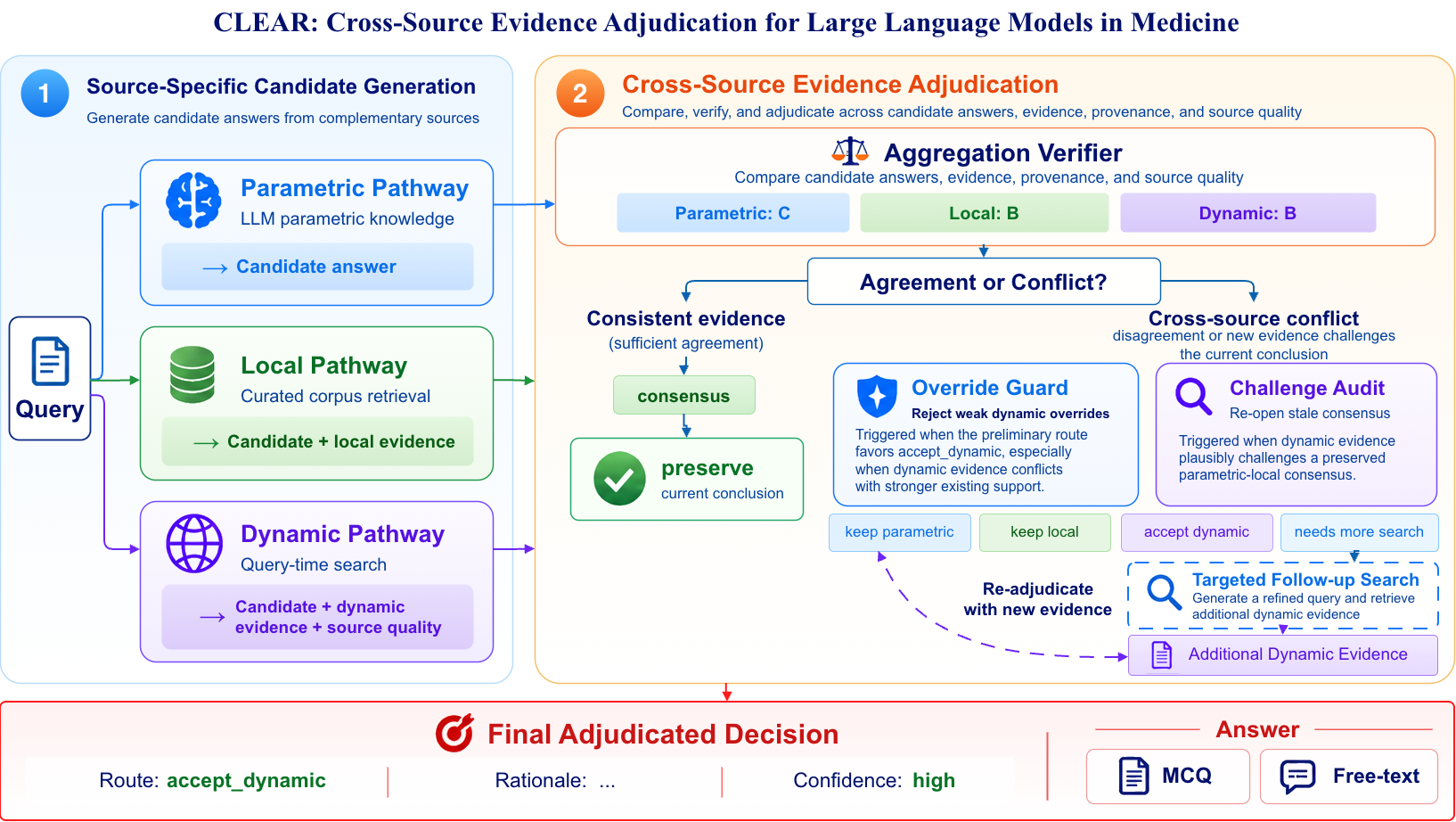}
    \caption{
    Overview of CLEAR. The framework independently generates candidate answers from parametric knowledge, locally curated evidence, and dynamically retrieved evidence; aggregates and adjudicates information across sources; and produces task-specific outputs.
    }
    \label{fig:framework}
\end{figure}

To address this challenge, we propose \textbf{CLEAR}, an agentic framework for cross-source evidence adjudication in LLMs in medicine. CLEAR generates source-specific candidate answers from parametric knowledge, locally curated evidence, and dynamically retrieved evidence and explicitly resolves their agreement and conflicts before producing the final response. CLEAR first constructs three parallel reasoning pathways, allowing each source to independently contribute a candidate answer and its supporting evidence. An aggregation verifier then reasons over the resulting evidence bundle, determines whether the sources agree, conflict, or remain insufficiently supported, and assigns an appropriate routing decision. When dynamic evidence challenges an agreement between the parametric and local pathways, an adjudication module further determines whether the new evidence is sufficiently relevant and reliable to justify revising the existing conclusion. This module combines an override guard, which prevents weak or misaligned dynamic evidence from inappropriately overturning a well-supported consensus, with a challenge audit, which allows credible new evidence to correct a consensus that may be outdated or incomplete. When the available evidence cannot resolve the conflict, the framework escalates the case to targeted follow-up search before making the final decision.

We systematically evaluate CLEAR on ten public benchmarks spanning medical question answering, clinical reasoning, and broader knowledge and reasoning tasks using representative commercial and open-weight backbones. CLEAR remains competitive in settings where Direct inference or conventional retrieval already performs strongly, while achieving substantial gains in several other settings. In particular, with Qwen-3.5-9B, the best CLEAR configurations improve performance by 13.69 percentage points on MedRBench and 11.30 percentage points on NEJM-QA over the strongest evaluated baselines. These findings suggest that effective use of external evidence requires not only retrieval, but also explicit adjudication of whether to preserve an existing conclusion, revise it, or seek additional evidence. We also publicly release the implementation code to support reproducibility and future research.
\section{Method}
\label{sec:method}

\subsection{Framework Overview}

Figure~\ref{fig:framework} illustrates CLEAR. Given a query $q$, such as a medical question, the framework considers three complementary sources of knowledge and evidence: (1) parametric knowledge encoded in the backbone model, (2) evidence retrieved from locally curated corpora, and (3) dynamic evidence acquired at inference time from external information sources, such as web search. Together, these sources capture a common practical setting in which a model must combine its internal knowledge with controlled local resources and newly acquired external information~\cite{wang2025bring,zhang2023large}.

CLEAR comprises three stages. First, a multi-path generation module independently constructs candidate answers from the parametric, local, and dynamic pathways. Maintaining separate pathways preserves the provenance of each candidate and makes cross-source agreement and disagreement observable. Second, an aggregation verifier assembles the candidate answers, retrieved evidence, and available source-quality information into a unified evidence bundle. It evaluates the support for each candidate and assigns a routing decision according to whether the sources agree, conflict, or remain insufficient to support a conclusion. Third, an adjudication module determines whether the current conclusion should be preserved, revised in light of dynamic evidence, or subjected to targeted follow-up search.

CLEAR is conservative by design. Dynamic evidence is not preferred solely because it is newly retrieved. When the parametric and local pathways agree, their consensus is preserved unless the dynamic pathway provides sufficiently relevant and reliable evidence that directly resolves the disagreement. Conversely, CLEAR does not assume that an existing consensus is necessarily correct. A challenge audit allows credible dynamic evidence to revise a conclusion that may be outdated, incomplete, or insufficiently sensitive to the specific context. When the available evidence cannot resolve the conflict, the system performs a bounded follow-up search before producing the final response. The individual components are described below.

\subsection{Source-Specific Candidate Generation}

Given the query $q$, CLEAR generates three source-specific candidate answers. Unless otherwise specified, all candidate-generation pathways use the same backbone model $\mathcal{M}$, thereby reducing differences attributable to model capability and allowing the pathways to be compared primarily with respect to the information available to each. Variants that introduce an additional model for evidence assessment are described separately in the experimental setup.

\paragraph{Parametric pathway.}
The parametric pathway generates an answer using only the knowledge encoded in the backbone model:

\begin{equation}
a_{\mathrm{param}}
=
\mathcal{M}(q).
\end{equation}

This pathway provides a retrieval-free reference and preserves the model's learned medical priors. It is particularly important when the query is already well supported by parametric knowledge and additional retrieval would be redundant or potentially distracting~\cite{shi2024enhancing,kim2025rethinking}.

\paragraph{Local-evidence pathway.}
Let $K_{\mathrm{local}}$ denote a locally indexed corpus or collection of corpora, and let $\mathcal{R}$ denote a retrieval function over this collection. The retrieved local evidence, its provenance, and the corresponding candidate answer are defined as

\begin{equation}
E_{\mathrm{local}},
P_{\mathrm{local}}
=
\mathcal{R}(q,K_{\mathrm{local}}),
\qquad
a_{\mathrm{local}}
=
\mathcal{M}(q,E_{\mathrm{local}}),
\end{equation}

where $P_{\mathrm{local}}$ records available provenance information associated with the retrieved local evidence, such as document identifiers or retrieval sources.

This pathway follows established RAG implementations~\cite{xiong2024benchmarking,liu2025improving}. The proposed framework is agnostic to the implementation of $\mathcal{R}$ and can therefore support sparse, dense, hybrid, or other retrieval approaches. The local pathway represents settings in which evidence is curated, controlled, domain-specific, or institution-specific, but potentially limited by the coverage and update time of the indexed corpus.

\paragraph{Dynamic-evidence pathway.}
The dynamic pathway acquires evidence from an external information space $K_{\mathrm{dynamic}}$ at inference time:

\begin{equation}
E_{\mathrm{dynamic}},
P_{\mathrm{dynamic}}
=
\mathrm{Search}(q,K_{\mathrm{dynamic}}),
\end{equation}

where $P_{\mathrm{dynamic}}$ records the provenance of the retrieved external evidence. The framework further assesses the quality of the retrieved evidence:

\begin{equation}
S_{\mathrm{dynamic}}
=
\mathrm{AssessQuality}
\left(
q,
E_{\mathrm{dynamic}},
P_{\mathrm{dynamic}}
\right),
\qquad
a_{\mathrm{dynamic}}
=
\mathcal{M}(q,E_{\mathrm{dynamic}}),
\end{equation}

where $S_{\mathrm{dynamic}}$ summarizes source-quality information derived from the retrieved evidence, such as assessments based on source authority, recency, and content type. Dynamic retrieval may be implemented using web search, literature search, external databases, or other query-time information tools.

Each pathway produces a source-specific candidate record:

\begin{equation}
c_i
=
(a_i,E_i,P_i),
\qquad
i
\in
\{
\mathrm{param},
\mathrm{local},
\mathrm{dynamic}
\},
\end{equation}

where $P_i$ records the provenance of the candidate and its supporting evidence when available. For the parametric pathway, $E_{\mathrm{param}}=\varnothing$ because its supporting knowledge is implicit in the model parameters, and $P_{\mathrm{param}}$ identifies the parametric pathway and backbone model. For the local and dynamic pathways, $E_i$ contains the retrieved evidence and $P_i$ records its retrieval provenance.

The resulting candidate set is

\begin{equation}
\mathcal{C}
=
\{
c_{\mathrm{param}},
c_{\mathrm{local}},
c_{\mathrm{dynamic}}
\}.
\end{equation}

The purpose of multi-path generation is not to assume that using more sources necessarily improves performance. Instead, it exposes source-specific conclusions and their supporting information so that the subsequent modules can explicitly evaluate agreement, disagreement, and evidential sufficiency.

\subsection{Cross-Source Aggregation and Adjudication}

\subsubsection{Aggregation Verifier}

The candidate pathways provide complementary answers, but they do not by themselves determine which answer should govern the final response. The aggregation verifier therefore constructs a unified evidence bundle:

\begin{equation}
B
=
\mathrm{Aggregate}
\left(
q,
\mathcal{C},
S_{\mathrm{dynamic}}
\right).
\end{equation}

The bundle $B$ contains the original query, the three candidate answers, the retrieved local and dynamic evidence, the provenance of each candidate, and the available source-quality information. The aggregation verifier $\mathcal{V}$ reasons over this bundle and returns a structured preliminary decision:

\begin{equation}
o_0
=
\mathcal{V}(B)
=
(\widehat{a}_0,z_0,j_0,\gamma_0),
\end{equation}

where $\widehat{a}_0$ is the preliminary answer, $z_0$ is the routing decision, $j_0$ is the supporting rationale, and $\gamma_0$ is a confidence signal associated with the decision. The confidence signal is not treated as a calibrated probability.

The routing decision belongs to the following set:

\begin{equation}
\begin{aligned}
\mathcal{Z}
= \{&
\texttt{consensus},
\texttt{keep\_parametric},
\texttt{keep\_local}, \\
&
\texttt{accept\_dynamic},
\texttt{needs\_more\_search}
\}.
\end{aligned}
\end{equation}

The routes have the following interpretations:

\begin{itemize}
    \item \texttt{consensus}: At least two pathways converge on the same conclusion, and the available supporting evidence does not reveal a material unresolved contradiction.

    \item \texttt{keep\_parametric}: The parametric answer is better supported than the alternatives, and the retrieved evidence does not justify revising it.

    \item \texttt{keep\_local}: The locally grounded answer is better supported by the curated evidence than the parametric or dynamic alternatives.

    \item \texttt{accept\_dynamic}: The dynamic answer is supported by sufficiently relevant and reliable external evidence and appears to justify revising the current conclusion.

    \item \texttt{needs\_more\_search}: The available evidence is insufficient, internally inconsistent, or unable to distinguish reliably among competing conclusions.
\end{itemize}

The aggregation verifier therefore goes beyond simple answer aggregation or majority voting. It jointly evaluates the provenance, relevance, and support associated with each candidate and determines whether the observed disagreement can be resolved using the current evidence bundle.

\subsubsection{Adjudication of Cross-Source Conflicts}

The adjudication module is invoked when the verifier identifies a meaningful conflict or when the preliminary route may require revising an existing conclusion. A particularly important conflict pattern is

\begin{equation}
a_{\mathrm{param}}
=
a_{\mathrm{local}}
\neq
a_{\mathrm{dynamic}},
\end{equation}

in which the parametric and local pathways agree while the dynamic pathway proposes a different conclusion. This pattern is inherently ambiguous. The dynamic pathway may have identified newer or more context-specific evidence, but it may also have retrieved irrelevant, unreliable, or question-misaligned information. The adjudication module therefore evaluates both the risk of an inappropriate override and the risk of preserving an outdated or incomplete consensus.

\paragraph{Override guard.}
The override guard evaluates cases in which the preliminary decision favors \texttt{accept\_dynamic}, particularly when the dynamic candidate conflicts with an agreement between the parametric and local pathways or when all three pathways disagree. The guard permits an override only when the dynamic evidence:

\begin{enumerate}
    \item directly addresses the query and the observed disagreement;
    \item originates from sufficiently credible and relevant sources;
    \item provides explicit support for the dynamic conclusion; and
    \item explains why the competing conclusion is less appropriate.
\end{enumerate}

If these conditions are not adequately supported, the dynamic candidate is not immediately accepted. The system either preserves the better-supported parametric or local conclusion or initiates targeted follow-up retrieval. This mechanism reduces the risk that weak, unreliable, or misaligned external evidence will overturn a stronger conclusion.

\paragraph{Challenge audit.}
The challenge audit addresses the opposite failure mode. A parametric--local consensus may itself be outdated, incomplete, or insensitive to the specific clinical context. Therefore, when the preliminary decision favors preserving the existing conclusion but the dynamic pathway raises a plausible and well-supported challenge, the audit explicitly re-examines the consensus.

The challenge is accepted only when the dynamic evidence satisfies the specified relevance and quality criteria, directly addresses the disagreement, and provides sufficient support for the revised conclusion. Otherwise, the original conclusion is preserved. The challenge audit therefore prevents agreement between the parametric and local pathways from being treated as automatically correct.

Together, the override guard and challenge audit balance two competing risks: allowing unreliable dynamic evidence to inappropriately overturn a well-supported conclusion, and allowing an outdated consensus to suppress credible and more current evidence.

\subsubsection{Targeted Follow-Up Search}

When the current evidence bundle cannot resolve the disagreement, the framework generates a targeted follow-up query conditioned on the unresolved conflict:

\begin{equation}
\widetilde{q}
=
\mathcal{Q}_{\mathrm{follow}}
(q,B,o_0),
\end{equation}

where $\mathcal{Q}_{\mathrm{follow}}$ identifies the specific evidential gap, competing claims, or clinical distinction requiring clarification. Additional evidence and its provenance are then acquired using the targeted query:

\begin{equation}
E_{\mathrm{follow}},
P_{\mathrm{follow}}
=
\mathrm{Search}
(\widetilde{q},K_{\mathrm{dynamic}}).
\end{equation}

The framework assesses the quality of this additional evidence as

\begin{equation}
S_{\mathrm{follow}}
=
\mathrm{AssessQuality}
\left(
\widetilde{q},
E_{\mathrm{follow}},
P_{\mathrm{follow}}
\right),
\end{equation}

and merges the new evidence with the existing evidence bundle:

\begin{equation}
B'
=
\mathrm{Merge}
\left(
B,
E_{\mathrm{follow}},
P_{\mathrm{follow}},
S_{\mathrm{follow}}
\right).
\end{equation}

The aggregation verifier and adjudication module are subsequently applied to the updated bundle $B'$. Follow-up retrieval is bounded rather than open-ended and is used only when the current evidence is insufficient to justify either preserving or revising the conclusion. Unlike unrestricted repeated search, the follow-up query is directed at the specific conflict identified during adjudication.

The adjudication process returns a structured decision object:

\begin{equation}
d
=
\left(
a^\star,
z^\star,
j^\star,
\gamma^\star,
B^\star
\right),
\end{equation}

where $a^\star$ is the selected answer, $z^\star$ is the final route, $j^\star$ is the adjudication rationale, $\gamma^\star$ is the final confidence signal, and $B^\star$ is the final evidence bundle after any follow-up search.

\subsection{Task-Specific Response Generation}

The cross-source adjudication procedure above is shared across different tasks. Once the structured decision $d$ has been obtained, a task-specific adapter converts the adjudicated conclusion and its supporting evidence into the output format required by the target task.

\paragraph{Free-text medical tasks.}
For an open-ended task, the final response is generated as

\begin{equation}
r_{\mathrm{text}}
=
\mathcal{G}_{\mathrm{text}}
(q,a^\star,j^\star,B^\star).
\end{equation}

The response generator does not independently repeat retrieval or overturn the adjudication result. Its role is to express the selected conclusion and supporting evidence in a medically grounded and task-appropriate form. Depending on the application, the response may include a diagnosis, treatment recommendation, clinical explanation, or patient-facing answer.

\paragraph{Multiple-choice tasks.}
For a multiple-choice task with an admissible option set $\mathcal{O}$, the task-specific adapter maps the adjudicated conclusion to one or more valid answer options:

\begin{equation}
\widehat{\mathcal{O}}
=
\mathcal{T}_{\mathrm{MCQ}}
(q,\mathcal{O},d),
\qquad
\widehat{\mathcal{O}}
\subseteq
\mathcal{O}.
\end{equation}

For single-answer tasks, $|\widehat{\mathcal{O}}|=1$; for multiple-answer tasks, $\widehat{\mathcal{O}}$ may contain more than one admissible option.

When the initial evidence is insufficient to distinguish among competing answer options, the targeted follow-up query may be conditioned on those options so that the search directly evaluates the relevant clinical distinctions. This option-aware retrieval occurs before the final adjudication decision $d$ is produced and therefore does not constitute a separate decision process.

After adjudication, a format gate converts the selected option or options into the representation required by the benchmark:

\begin{equation}
r_{\mathrm{MCQ}}
=
\mathrm{FormatGate}
(\widehat{\mathcal{O}},\mathcal{O}).
\end{equation}

The format gate does not alter the medical conclusion. It only ensures that the output contains the appropriate number and representation of admissible options, thereby supporting both single-answer and multiple-answer tasks.

\subsection{Implementation Details}

We instantiate CLEAR with GPT-o3-mini and Qwen-3.5-9B as representative commercial and open-weight backbones, respectively. In the primary setting, the same backbone is used throughout candidate generation, source-quality assessment, aggregation verification, adjudication, and final response generation. Full model configurations and prompting details are provided in Appendix~\ref{app:details}.

For candidate generation, the parametric pathway directly queries the backbone without external retrieval. The local pathway retrieves evidence from MedCorp~\cite{xiong2024benchmarking}, following the same corpus setting as the MedRAG baselines, using BM25 with the top $k=16$ retrieved snippets. The dynamic pathway obtains external evidence through web search. The primary experiments use GPT-4o's web-search capability for dynamic evidence acquisition. As a secondary implementation-diversity experiment, we additionally instantiate the dynamic pathway using Tavily on the NEJM-QA Internal Medicine subset (Appendix~\ref{app:details}), providing preliminary evidence that CLEAR can be implemented with an alternative web-search backend.

Retrieved dynamic evidence is assigned categorical source-quality labels (high, medium, or low) based on source authority, recency, and content type. These labels constitute $S_{\mathrm{dynamic}}$ and are provided to the aggregation verifier and adjudication modules together with the candidate answers, retrieved evidence, and provenance information. The aggregation verifier produces a preliminary answer and routing decision. When dynamic evidence challenges another pathway, the adjudication module applies two complementary mechanisms: an \emph{override guard}, which verifies a verifier-proposed dynamic override before accepting it, and a \emph{challenge audit}, which allows rejected dynamic evidence to challenge an existing parametric--local consensus when sufficiently strong supporting evidence is available. Detailed triggering conditions and decision rules are provided in Appendix~\ref{app:details}.

Finally, task-specific adapters convert the adjudicated conclusion into benchmark-compatible outputs for multiple-choice tasks or evidence-grounded responses for free-text tasks, without introducing additional retrieval or modifying the adjudication result.

\subsection{Data and Evaluation}

We evaluate CLEAR on ten public reasoning benchmarks, including eight medical question-answering or clinical reasoning benchmarks and two broader knowledge and reasoning benchmarks. The evaluation suite was selected to balance three considerations: comparability with prior work, task difficulty, and diversity in answer format. Collectively, the benchmarks span multiple-choice and free-text tasks as well as general and specialty-level medical knowledge.

Due to space constraints, we focus the main-text analysis on MedQA, PubMedQA, NEJM-QA, MedRBench, and HealthBench. We additionally evaluate the framework on MedBullets~\cite{chen2025benchmarking}, MedExQA~\cite{kim2024medexqa}, AfriMed-QA~\cite{nimo2025afrimed}, MMLU~\cite{hendrycks2020measuring}, and MMLU-Pro~\cite{wang2024mmlu}. The magnitude of improvement varies across datasets and backbone models; complete results on the five additional benchmarks are provided in Appendix~\ref{sec:app_additional_results}.

\paragraph{Multiple-choice benchmarks.}
We evaluate multiple-choice question answering on MedQA, PubMedQA, and NEJM-QA. MedQA~\cite{jin2021disease} is a widely used benchmark based on United States Medical Licensing Examination-style questions. Its English test set contains 1,273 questions, each with four answer options, and primarily evaluates professional medical knowledge and exam-style clinical reasoning.

PubMedQA~\cite{jin2019pubmedqa} is constructed from biomedical research questions and corresponding PubMed abstracts. Each question is answered as \textit{yes}, \textit{no}, or \textit{maybe} based on the associated abstract, making the benchmark more focused on interpreting biomedical evidence than on isolated factual recall. We use its 500 expert-labeled test questions.

NEJM-QA~\cite{wang2025baichuan} is constructed from official board residency examinations released through the NEJM-AI collection. It contains 655 questions spanning five specialties: general surgery, internal medicine, psychiatry, pediatrics, and obstetrics and gynecology. Relative to MedQA and PubMedQA, NEJM-QA emphasizes more specialized medical knowledge and provides a setting with multiple clinically plausible alternatives. Some NEJM-QA questions admit multiple correct answers; for these questions, a prediction is considered correct only when the complete predicted set exactly matches the reference answer set.

We retain MedQA and PubMedQA not because they are expected to benefit most from retrieval, but because they represent standard evaluation settings in which strong parametric models and conventional local retrieval already perform well. These benchmarks therefore test whether the proposed framework can preserve performance when additional external evidence is unnecessary or potentially distracting.

\paragraph{Free-text benchmarks.}
We also evaluate CLEAR on free-text benchmarks MedRBench and HealthBench. MedRBench~\cite{qiu2025quantifying} contains 1,453 structured patient cases with reference reasoning derived from open-access clinical case reports. The benchmark spans 13 body systems and 10 specialties and includes cases involving rare diseases. Following the benchmark protocol, we evaluate the diagnosis and treatment stages using its automated reasoning evaluator.

HealthBench~\cite{arora2025healthbench} contains 5,000 realistic multi-turn health conversations graded using conversation-specific rubrics developed by 262 physicians. Rather than requiring selection from fixed answer options, HealthBench evaluates open-ended response quality across criteria including clinical accuracy, completeness, safety, and communication. We use the HealthBench Hard subset, comprising 1,000 queries.

\paragraph{Additional benchmarks.}
To assess robustness across a broader range of reasoning settings, we additionally evaluate CLEAR on five benchmarks. MedBullets contains USMLE Step 2 and Step 3-style multiple-choice questions with expert-written explanations. MedExQA covers multiple medical specialties and provides multiple reference explanations for each question. AfriMed-QA is a Pan-African, multi-specialty medical question-answering benchmark constructed across medical institutions in multiple African countries. MMLU evaluates multitask knowledge across a broad range of academic and professional subjects, while MMLU-Pro provides a more challenging reasoning-oriented extension with a larger answer space. Detailed dataset statistics and complete results are provided in Appendix~\ref{sec:app_additional_results}.

\paragraph{Evaluation metrics.}
Because the benchmarks differ in answer format and evaluation objective, we use task-specific metrics. For single-answer multiple-choice questions, we report accuracy based on the final selected option. For multiple-answer NEJM-QA questions, we use exact-set accuracy, requiring the predicted option set to match the complete reference set. For MedRBench, we follow the official evaluation protocol and report the automated reasoning scores for the diagnosis and treatment stages, together with their aggregate result. For HealthBench, we report the physician-authored rubric score, which measures the extent to which each response satisfies its conversation-specific clinical criteria. All benchmark evaluators are used only for scoring and do not participate in candidate generation, evidence adjudication, or final response generation.

\subsection{Baselines and Framework Variants}

We evaluate all methods using the same answer-generation backbones to control for differences in the underlying model when comparing retrieval and evidence-adjudication strategies. Specifically, we use o3-mini as a commercial backbone and Qwen-3.5-9B as an open-weight backbone. This pairing allows us to examine whether the proposed framework remains effective across models with different capability levels and accessibility.

\paragraph{Direct inference.}
The \textbf{Direct} baseline answers each query using only the parametric knowledge encoded in the backbone model:

\begin{equation}
a_{\mathrm{direct}}
=
\mathcal{M}(q).
\end{equation}

This retrieval-free setting serves as the primary reference for assessing whether locally retrieved evidence, dynamically acquired evidence, and cross-source adjudication improve upon parametric reasoning alone.

\paragraph{Retrieval-augmented generation.}
We use MedRAG~\cite{xiong2024benchmarking} as the retrieval-augmented baseline because it provides an established framework for medical question answering and supports controlled evaluation with different retrieval methods. We evaluate MedRAG using two retrieval configurations: \textbf{MedRAG (BM25)}, based on sparse lexical retrieval, and \textbf{MedRAG (MedCPT)}, based on dense biomedical retrieval. Both configurations retrieve from the same local corpus and use the same answer-generation backbone as the corresponding Direct and proposed-method settings.

These two configurations help assess whether improvements extend beyond those obtained by changing the local retrieval method. Conceptually, MedRAG represents the local-evidence pathway,

\begin{equation}
a_{\mathrm{local}}
=
\mathcal{M}
\left(
q,
\mathcal{R}(q,K_{\mathrm{local}})
\right),
\end{equation}

whereas CLEAR jointly considers parametric, local, and dynamic pathways and explicitly adjudicates their agreement and conflicts.

\paragraph{Proposed framework.}
We evaluate three configurations of the proposed framework:

\begin{itemize}
    \item \textbf{CLEAR (default)} is the primary configuration. It includes source-specific candidate generation, the aggregation verifier, the override guard, the challenge audit, bounded follow-up search, and task-specific response generation.

    \item \textbf{CLEAR (dual models)} follows the same adjudication procedure but introduces a second model to assess the relevance and quality of dynamically retrieved evidence. This configuration evaluates whether model diversity during evidence assessment improves cross-source adjudication.

    \item \textbf{CLEAR (QPIS)} extends the primary configuration with Query Planning and Iterative Search. It decomposes complex queries into targeted subqueries and iteratively acquires additional evidence when the current evidence bundle remains insufficient. This configuration is designed for questions requiring broader evidence coverage or multi-step information seeking.
\end{itemize}

The three configurations share the same cross-source adjudication architecture and differ primarily in how dynamic evidence is acquired or assessed, including the use of an additional assessment model or iterative query planning. Unless otherwise stated, \textbf{CLEAR (default)} is treated as the primary proposed method, whereas the dual-model and QPIS configurations are evaluated as extensions.
\section{Results}
\label{sec:results}

In this section, we systematically present the main experimental results of CLEAR. We evaluate the framework separately on multiple-choice question (MCQ) and free-text benchmarks and further conduct ablation studies on its two key mechanisms. To keep the main text focused, we report the primary results and analyses here, while additional experimental results are provided in Section~\ref{sec:app_additional_results}.

\subsection{Results on MCQ Datasets}

As shown in Table~\ref{tab:common-mcq-results}, \textbf{CLEAR} achieves competitive results across the MCQ datasets.


\begin{table*}[t]
\centering
\caption{Results on Common MCQ Datasets}
\label{tab:common-mcq-results}
\resizebox{\textwidth}{!}{
\small
\setlength{\tabcolsep}{5pt}
\renewcommand{\arraystretch}{1.08}
\begin{tabular}{llccc}
\toprule
Method & Backbone & MedQA & PubMedQA & NEJM Overall \\
\midrule

\multicolumn{5}{l}{\textbf{GPT-based Model}} \\
\midrule
Direct
& o3-mini
& 92.38 & 79.40 & 84.12 \\

MedRAG (BM25 Setting)
& o3-mini
& 92.14 & \textbf{82.00} & 80.76 \\

MedRAG (MedCPT Setting)
& o3-mini
& 91.59 & 80.60 & 80.61 \\

CLEAR
& o3-mini
& 92.69 & 81.60 & \textbf{85.04} \\

CLEAR (dual models)
& o3-mini
& 92.14 & 81.80 & 84.73 \\

CLEAR (Query Planning + Iterative Search)
& o3-mini
& \textbf{92.85} & 80.60 & 84.43 \\

\midrule
\multicolumn{5}{l}{\textbf{Qwen-based Model}} \\
\midrule
Direct
& Qwen-3.5-9B
& 72.19 & 78.40 & 66.56 \\

MedRAG (BM25 Setting)
& Qwen-3.5-9B
& 73.84 & \textbf{80.00} & 66.87 \\

MedRAG (MedCPT Setting)
& Qwen-3.5-9B
& 72.66 & 79.40 & 66.87 \\

CLEAR
& Qwen-3.5-9B
& 79.97 & 79.20 & 73.74 \\

CLEAR (dual models)
& Qwen-3.5-9B
& \textbf{83.58} & 79.60 & 75.42 \\

CLEAR (Query Planning + Iterative Search)
& Qwen-3.5-9B
& 83.19 & 79.40 & \textbf{78.17} \\

\bottomrule
\end{tabular}
}
\end{table*}

In detail, on MedQA, the \textbf{CLEAR variants} achieve performance comparable to or better than the Direct and MedRAG baselines across both the o3-mini and Qwen backbones. On PubMedQA, although the CLEAR variants do not surpass MedRAG under the BM25 setting, their performance remains close. Together, these results show that CLEAR remains competitive with the Direct and MedRAG baselines on commonly used medical QA benchmarks.

NEJM-QA targets more specialized medical knowledge than MedQA and PubMedQA. For overall NEJM-QA accuracy, all \textbf{CLEAR} variants outperform the evaluated baselines. This advantage is particularly pronounced with the Qwen backbone, where the best CLEAR configuration improves accuracy by 11.30 percentage points over the strongest baseline.


\begin{table*}[t]
\centering
\caption{NEJM Detail Results}
\label{tab:nejm-detail-results}
\small
\setlength{\tabcolsep}{4.5pt}
\renewcommand{\arraystretch}{1.08}
\resizebox{\textwidth}{!}{
\begin{tabular}{llccccc}
\toprule
Method & Backbone & General Surgery & Internal Medicine & Psychiatry & Pediatrics & Obgyn \\
\midrule

\multicolumn{7}{l}{\textbf{GPT-based Model}} \\
\midrule
Direct
& o3-mini
& 85.82 & 87.30 & 83.33 & 83.84 & 80.58 \\

MedRAG (BM25 Setting)
& o3-mini
& 80.85 & 80.16 & 79.33 & 86.87 & 78.42 \\

MedRAG (MedCPT Setting)
& o3-mini
& 78.72 & 84.92 & 82.67 & 81.82 & 75.54 \\

CLEAR
& o3-mini
& \textbf{87.94} & 83.33 & \textbf{85.33} & \textbf{87.88} & 81.29 \\

CLEAR (dual models)
& o3-mini
& 87.23 & 82.54 & 84.00 & 86.87 & \textbf{83.45} \\

CLEAR (Query Planning + Iterative Search)
& o3-mini
& 85.11 & \textbf{88.89} & 82.67 & 86.87 & 79.86 \\

\midrule
\multicolumn{7}{l}{\textbf{Qwen-based Model}} \\
\midrule
Direct
& Qwen-3.5-9B
& 62.41 & 64.29 & 77.33 & 62.63 & 64.03 \\

MedRAG (BM25 Setting)
& Qwen-3.5-9B
& 67.38 & 65.87 & 73.33 & 66.67 & 60.43 \\

MedRAG (MedCPT Setting)
& Qwen-3.5-9B
& 63.83 & 65.08 & 75.33 & 64.65 & 64.03 \\

CLEAR
& Qwen-3.5-9B
& 71.63 & 75.40 & 81.33 & 71.72 & 67.63 \\

CLEAR (dual models)
& Qwen-3.5-9B
& \textbf{75.89} & 74.60 & 81.33 & 75.76 & 69.06 \\

CLEAR (Query Planning + Iterative Search)
& Qwen-3.5-9B
& 75.18 & \textbf{79.37} & \textbf{84.00} & \textbf{78.79} & \textbf{73.38} \\

\bottomrule
\end{tabular}
}
\end{table*}

\begin{figure}[htbp]
    \centering
    \includegraphics[width=\textwidth]{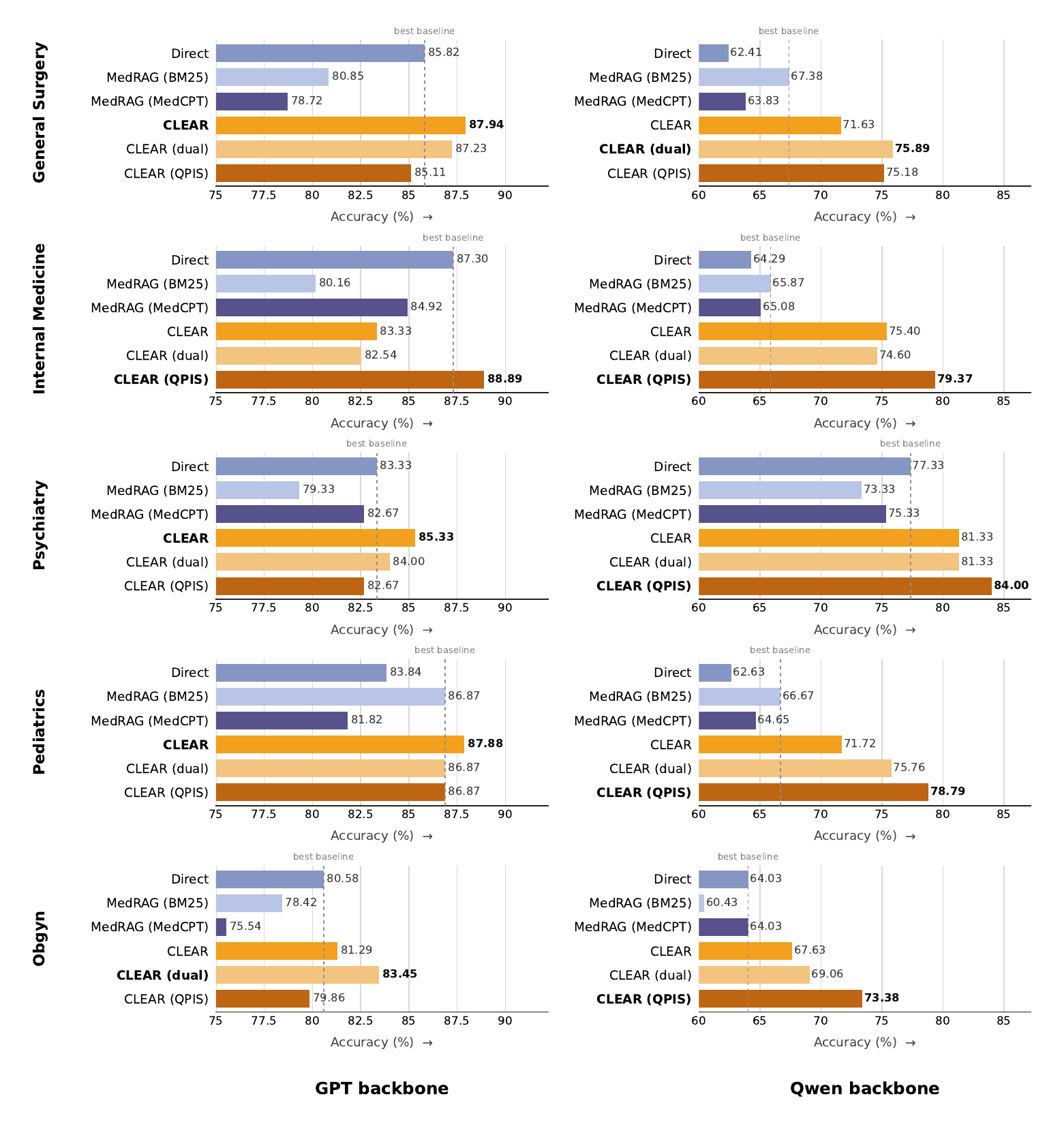}
    \caption{Specialty-level results on NEJM-QA. At least one CLEAR configuration achieves the best performance in each evaluated specialty across both backbone families.}
    \label{fig:mcq_nejm}
\end{figure}

As shown in Figure~\ref{fig:mcq_nejm}, we further report specialty-level results on NEJM-QA. Across the two backbones, at least one \textbf{CLEAR} configuration achieves the highest performance in each evaluated specialty. The gains are particularly consistent with the Qwen backbone.

\begin{figure}[htbp]
    \centering
    \includegraphics[width=\textwidth]{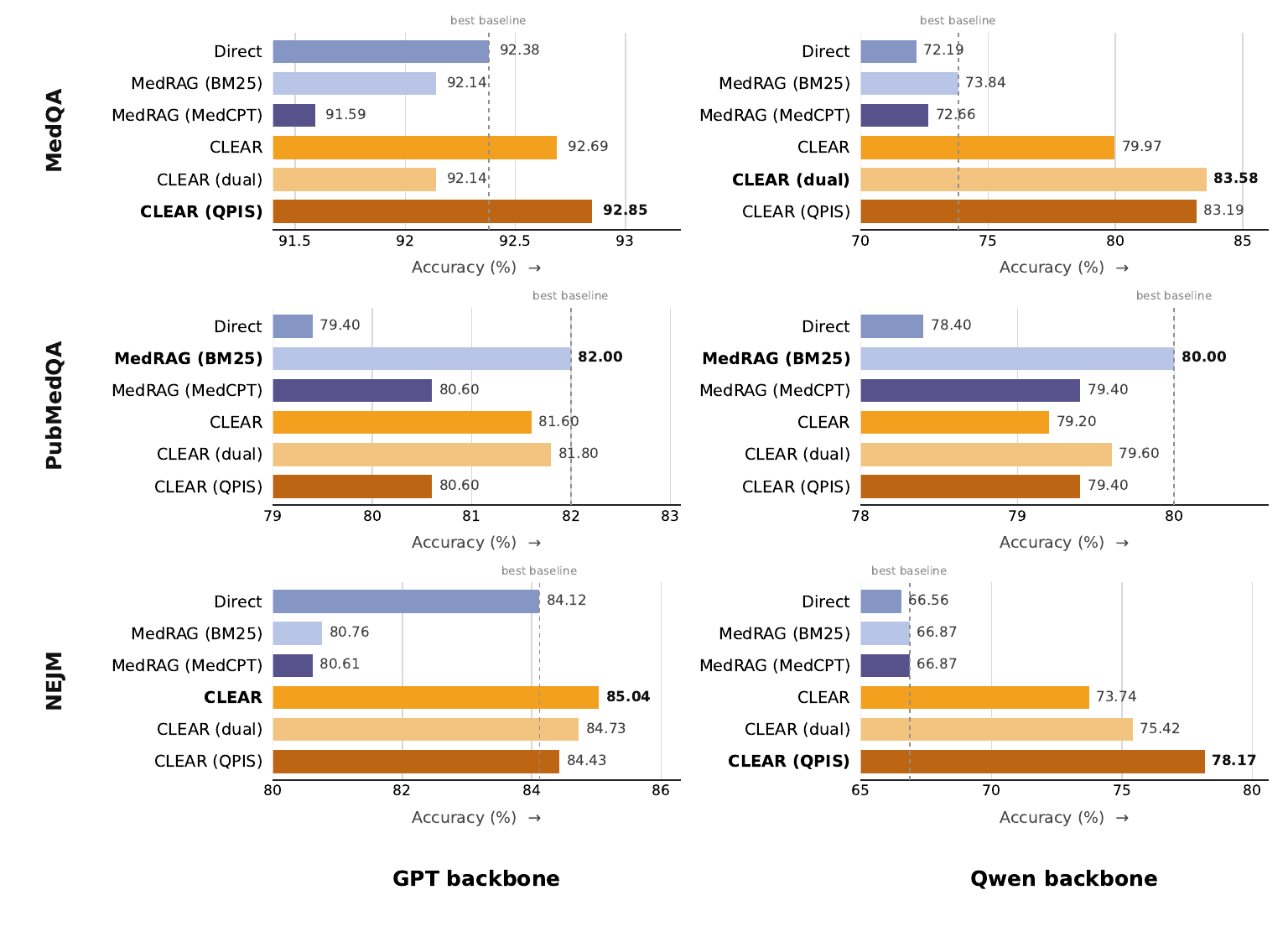}
    \caption{Performance on MedQA, PubMedQA, and NEJM-QA. CLEAR remains competitive on MedQA and PubMedQA while showing larger gains on NEJM-QA, especially with the Qwen backbone.}
    \label{fig:r_fig_mcq}
\end{figure}

In contrast to the results on MedQA and PubMedQA, the MedRAG baseline, which retrieves from a local corpus, shows lower performance on NEJM-QA under both the BM25 and MedCPT retriever settings. With o3-mini, MedRAG (BM25) and MedRAG (MedCPT) achieve 80.76\% and 80.61\%, respectively, compared with 84.12\% for Direct inference. A similar pattern is observed with Qwen-3.5-9B.

\subsection{Results on Free-Text Datasets}

To assess performance beyond multiple-choice question answering, we further evaluate CLEAR on MedRBench and HealthBench-Hard.

Table~3 reports results on the free-text benchmarks. On MedRBench, CLEAR scores 82.38 Overall with gpt-o3-mini (82.34 on Diagnosis, 82.46 on Treatment) and 74.74 with qwen-3.5-9B (76.70, 70.97). QPIS variant scores 79.77 and 73.30 Overall on the two backbones. Direct inference scores 71.30 and 61.05 Overall; MedRAG scores 71.03 (BM25) and 65.45 (MedCPT) with gpt-o3-mini, and 58.22 under both settings with qwen-3.5-9B.

HealthBench shows a different pattern from MedRBench. With gpt-o3-mini, CLEAR scores 0.2422 against 0.3334 for Direct inference, and falls below both MedRAG settings, while QPIS scores 0.3334. With qwen-3.5-9B, CLEAR scores 0.3444 against 0.3179 for Direct inference, and QPIS scores 0.3179. Both MedRAG settings score below Direct inference under both backbones. We discuss these backbone- and task-dependent differences in Section~\ref{sec:discussion}.



\begin{table*}[t]
\centering
\caption{Results on Free-text Benchmarks}
\label{tab:free-text-benchmark-results}
\resizebox{\textwidth}{!}{
\small
\setlength{\tabcolsep}{5pt}
\renewcommand{\arraystretch}{1.08}
\begin{tabular}{llcccc}
\toprule
Method & Backbone & MedRBench Overall & MedRBench D & MedRBench T & HealthBench \\
\midrule

\multicolumn{6}{l}{\textbf{GPT-based Model}} \\
\midrule
Direct
& o3-mini
& 71.30 & 71.06 & 71.77 & 0.3334 \\

MedRAG (BM25 Setting)
& o3-mini
& 71.03 & 69.70 & 73.59 & 0.3051 \\

MedRAG (MedCPT Setting)
& o3-mini
& 65.45 & 62.70 & 70.77 & 0.2997 \\

CLEAR
& o3-mini
& \textbf{82.38} & \textbf{82.34} & \textbf{82.46} & 0.2422 \\

CLEAR (Query Planning + Iterative Search)
& o3-mini
& 79.77 & 80.46 & 78.43 & \textbf{0.3334} \\

\midrule
\multicolumn{6}{l}{\textbf{Qwen-based Model}} \\
\midrule
Direct
& Qwen-3.5-9B
& 61.05 & 59.35 & 64.31 & 0.3179 \\

MedRAG (BM25 Setting)
& Qwen-3.5-9B
& 58.22 & 57.58 & 59.48 & 0.2588 \\

MedRAG (MedCPT Setting)
& Qwen-3.5-9B
& 58.22 & 57.58 & 59.48 & 0.2434 \\

CLEAR
& Qwen-3.5-9B
& \textbf{74.74} & \textbf{76.70} & \textbf{70.97} & \textbf{0.3444} \\

CLEAR (Query Planning + Iterative Search)
& Qwen-3.5-9B
& 73.30 & 76.18 & 67.74 & 0.3179 \\

\bottomrule
\end{tabular}
}
\end{table*}

\subsection{Ablation Studies}

We conduct two ablation studies to examine cross-source adjudication and the targeted follow-up search mechanism.

First, we compare different evidence configurations on the o3-mini backbone, as reported in Table~\ref{tab:abl-sources}. Direct uses parametric knowledge only, BM25 uses local retrieval, Online-only uses dynamic retrieval, and CLEAR combines all three sources through the full framework. Performance varies across datasets. On MedQA, CLEAR achieves the highest score of 92.69, compared with 92.38 for Direct, 92.14 for BM25, and 87.82 for Online-only. On PubMedQA, BM25 achieves the highest score of 82.00, followed by CLEAR at 81.60. On NEJM-QA, CLEAR obtains the highest score of 85.04, compared with 84.12 for Direct, 80.76 for BM25, and 78.32 for Online-only. Aggregated across the three datasets, CLEAR achieves the highest overall score of 88.34.

Second, we evaluate the effect of targeted follow-up search on the Qwen-3.5-9B backbone. As shown in Table~\ref{tab:abl-second-search}, CLEAR improves the overall score from 77.22 without follow-up search to 78.13 with follow-up search, corresponding to an absolute improvement of 0.91 percentage points. The improvement is 1.02 percentage points on MedQA and 1.37 percentage points on NEJM-QA, while the PubMedQA score remains unchanged at 79.20. Across the three datasets, follow-up search is triggered 123 times, including 87 times on MedQA, 4 times on PubMedQA, and 32 times on NEJM-QA. Among these triggered searches, 12 result in an accepted revision, corresponding to an acceptance rate of 9.8\%.

Overall, the ablation studies show that the full CLEAR framework achieves stronger aggregate performance than any individual evidence-source setting, while targeted follow-up search provides additional gains on a subset of unresolved cases.

\begin{table}[t]
\centering
\caption{Ablation Study I: Comparison across direct, BM25, online-only, and ours settings.}
\label{tab:abl-sources}
\begin{tabular}{lcccc}
\toprule
Dataset & Direct & BM25 & Online-only & CLEAR \\
\midrule
MedQA    & 92.38 & 92.14 & 87.82 & \textbf{92.69} \\
PubMedQA & 79.40 & \textbf{82.00} & 77.80 & 81.60 \\
NEJM-QA    & 84.12 & 80.76 & 78.32 & \textbf{85.04} \\
All      & 87.48 & 86.99 & 83.20 & \textbf{88.34} \\
\bottomrule
\end{tabular}
\end{table}
\begin{table}[t]
\centering
\caption{Ablation Study II: Impact and usage statistics of targeted follow-up searc}
\label{tab:abl-second-search}
\begin{tabular}{lcccc}
\toprule
Setting & MedQA & PubMedQA & NEJM-QA & All \\
\midrule
\multicolumn{5}{l}{\textit{Performance}} \\
CLEAR 
& \textbf{79.97} 
& \textbf{79.20} 
& \textbf{73.74} 
& \textbf{78.13} \\
w/o targeted follow-up search 
& 78.95 
& \textbf{79.20} 
& 72.37 
& 77.22 \\
Difference 
& +1.02 
& +0.00 
& +1.37 
& +0.91 \\
\midrule
\multicolumn{5}{l}{\textit{Second-search statistics}} \\
targeted follow-up search count 
& 87 
& 4 
& 32 
& 123 \\
targeted follow-up search accepted 
& 8 
& 1 
& 3 
& 12 \\
\bottomrule
\end{tabular}
\end{table}

\subsection{Result Interpretation and Manual Verification}

Because all benchmarks evaluated in this study are publicly available, web- or corpus-based retrieval may potentially recover either benchmark instances themselves or the upstream documents from which some benchmark items were constructed. We therefore manually audit every retrieval trace produced during evaluation and explicitly distinguish two cases.

\textbf{Direct benchmark-item leakage} refers to retrieval of the benchmark instance itself, such as an exact or near-exact match to the question stem together with its answer choices or reference answer. We detect such cases by checking for (1) contiguous question-stem matches, (2) high-overlap matches to the benchmark question, and (3) verbatim recovery of benchmark-specific answer options or reference answers.

\textbf{Upstream-source retrieval}, by contrast, refers to retrieval of the original publication, abstract, or clinical case from which a benchmark item was derived, without recovering the benchmark question, answer options, or reference answer. We identify such cases using source metadata and identifiers, including article titles, PubMed/PMC identifiers, and other benchmark-provided provenance information. The same source-identifier checks are also applied to the MedRAG baselines.

As shown in Table~\ref{tab:leakage}, we observe no direct benchmark-item leakage for any evaluated benchmark under either backbone: none of the retrieved documents contains the benchmark question-answer instance itself. This indicates that we did not identify direct recovery of benchmark items in the audited retrieval traces.

We next examine upstream-source retrieval. No upstream-source matches are observed for MedQA, NEJM-QA, or HealthBench. For PubMedQA, the source document is retrieved for 42.4\% and 42.6\% of examples with the GPT- and Qwen-based configurations, respectively. For MedRBench, the corresponding rates are 60.4\% and 62.5\% on the diagnosis subset and 75.0\% and 77.6\% on the treatment subset. The similar rates across backbones are consistent with upstream-source retrieval being driven primarily by the relationship between benchmark provenance and the retrieval space rather than by backbone-specific behavior.

Importantly, these upstream-source matches do not contain the benchmark items themselves. Instead, they correspond to titles, abstracts, or publicly available clinical cases from PubMed and PMC that served as source material for benchmark construction. Recovering such documents is an expected behavior of an effective retrieval system, whose objective is to identify documents that are highly relevant to a given query. For benchmarks derived from publicly indexed literature, the original source document may therefore be retrieved even when the derived benchmark question, answer options, and reference answer are not present.

This behavior is also observed in the MedRAG baselines. As shown in Table~\ref{tab:leakage_baseline}, on PubMedQA, BM25 retrieves the source abstract for 55.0\% of examples and MedCPT for 53.4\%, compared with 42.4\% for the CLEAR web-search configuration. The source abstract is ranked first for 55.0\% and 50.8\% of examples with BM25 and MedCPT, respectively. These results indicate that upstream-source retrieval is not unique to CLEAR and should be distinguished from direct benchmark-item leakage.

\begin{table}[htbp]
\centering
\caption{
Manual retrieval audit across benchmarks.
Direct benchmark-item leakage denotes retrieval of the benchmark question-answer instance itself, whereas upstream-source retrieval denotes recovery of the original source document used to construct the benchmark item without recovering the benchmark question or answer.
}
\label{tab:leakage}
\resizebox{\linewidth}{!}{
\begin{tabular}{lrrrrr}
\toprule
Benchmark & Items &
\multicolumn{2}{c}{Direct benchmark-item leakage} &
\multicolumn{2}{c}{Upstream-source retrieval} \\
\cmidrule(lr){3-4}
\cmidrule(lr){5-6}
& & Hits & Rate (\%) & Hits & Rate (\%) \\
\midrule

\multicolumn{6}{l}{\textit{GPT-based Model}} \\
MedQA & 1{,}273 & 0 & 0.0 & 0 & 0.0 \\
PubMedQA & 500 & 0 & 0.0 & 212 & 42.4 \\
NEJM-QA & 655 & 0 & 0.0 & 0 & 0.0 \\
HealthBench & 1{,}000 & 0 & 0.0 & 0 & 0.0 \\
MedRBench (diagnosis) & 957 & 0 & 0.0 & 578 & 60.4 \\
MedRBench (treatment) & 496 & 0 & 0.0 & 372 & 75.0 \\

\midrule

\multicolumn{6}{l}{\textit{Qwen-based Model}} \\
MedQA & 1{,}273 & 0 & 0.0 & 0 & 0.0 \\
PubMedQA & 500 & 0 & 0.0 & 213 & 42.6 \\
NEJM-QA & 655 & 0 & 0.0 & 0 & 0.0 \\
HealthBench & 1{,}000 & 0 & 0.0 & 0 & 0.0 \\
MedRBench (diagnosis) & 957 & 0 & 0.0 & 598 & 62.5 \\
MedRBench (treatment) & 496 & 0 & 0.0 & 385 & 77.6 \\
\bottomrule
\end{tabular}
}
\end{table}

\begin{table}[htbp]
\centering
\caption{
Upstream-source retrieval on PubMedQA. Recovery of upstream source documents is observed with both CLEAR web search and conventional MedRAG retrieval.
}
\label{tab:leakage_baseline}

\begin{tabular}{lrrrr}
\toprule
Retriever & Items & Source Retrieved & Rate (\%) & Top-1 (\%) \\
\midrule
CLEAR (web search) & 500 & 212 & 42.4 & --- \\
MedRAG (BM25)       & 500 & 275 & 55.0 & 55.0 \\
MedRAG (MedCPT)     & 500 & 267 & 53.4 & 50.8 \\
\bottomrule
\end{tabular}
\end{table}
\section{Discussion}
\label{sec:discussion}

\subsection{Main findings}

\paragraph{External evidence provides greater benefit when parametric knowledge provides weaker support.}
A main finding of this study is that the value of external evidence depends on how well the parametric knowledge of the backbone supports the task. When parametric knowledge is already strong, additional retrieval may provide limited benefit and can even degrade performance. For example, with o3-mini, Direct inference already achieves 92.38\% accuracy on MedQA and a rubric score of 0.3334 on HealthBench. CLEAR largely preserves performance on MedQA, whereas additional retrieved evidence does not improve HealthBench and, under the default configuration, reduces performance. These results suggest that retrieval should not be treated as uniformly beneficial when the model already provides a strong answer from its parametric knowledge.

In contrast, larger gains are observed in settings where Direct inference provides a weaker baseline. This pattern is particularly apparent with the open-weight Qwen-3.5-9B backbone. On MedQA, Direct inference achieves 72.19\% accuracy and the strongest conventional baseline reaches 73.84\%, whereas the best CLEAR configuration reaches 83.58\%, representing a 9.74-percentage-point improvement over the strongest baseline. On NEJM-QA, the strongest conventional baseline achieves 66.87\%, compared with 78.17\% for the best CLEAR configuration, a gain of 11.30 percentage points. A similar pattern is observed on MedRBench, where CLEAR improves the overall score from 61.05 with Direct inference to 74.74. Importantly, this benefit is not restricted to the open-weight backbone: with o3-mini, CLEAR improves MedRBench from 71.30 to 82.38. Thus, the benefit of external evidence appears to be task-dependent as well as model-dependent; even a strong backbone may benefit substantially in settings where parametric knowledge alone provides limited support. These observations are consistent with prior studies showing that the utility of retrieval depends on both the sufficiency of parametric knowledge and the quality of retrieved evidence~\cite{kim2025rethinking,wang2025astute}.

\paragraph{Retrieved evidence should not be trusted by default.}
More importantly, the need for additional knowledge does not imply that retrieved evidence should be trusted by default. The source-ablation results illustrate this distinction. With o3-mini, using dynamic evidence alone reduces accuracy from 92.38\% to 87.82\% on MedQA and from 84.12\% to 78.32\% on NEJM-QA. Local retrieval is also not uniformly beneficial: MedRAG improves performance on PubMedQA but reduces performance relative to Direct inference on several other benchmarks. These findings are consistent with prior evidence that retrieval may introduce irrelevant, misleading, or conflicting information and that models may fail to use retrieved evidence appropriately even when relevant information is available~\cite{xie2025retrieval,wong2025retrieval}. External evidence can therefore compensate for limitations in parametric knowledge, but it can simultaneously introduce a new source of uncertainty.

\paragraph{Cross-source adjudication provides a preliminary approach to evidence management.}
A third main finding is that explicit cross-source adjudication provides a preliminary approach to addressing these challenges. CLEAR maintains parametric, local, and dynamic evidence as separate candidate pathways rather than directly merging them into a single context. It then uses an aggregation verifier to identify agreement and disagreement among the sources, an override guard and challenge audit to determine whether a competing conclusion should revise the existing answer, and targeted follow-up search when the available evidence remains insufficient to resolve the conflict.

Across the ten evaluated benchmarks, CLEAR remains competitive across both multiple-choice and free-text settings, while achieving substantial gains in several settings where parametric or locally retrieved knowledge alone provides weaker performance. The gains are particularly consistent with the open-weight backbone. On the three primary MCQ benchmarks, the best CLEAR configurations improve MedQA by 9.74 percentage points and NEJM-QA by 11.30 percentage points over the strongest conventional baselines, while remaining close to the best result on PubMedQA. On MedRBench, the default CLEAR improves the overall score from 61.05 with Direct inference to 74.74 with Qwen-3.5-9B, and from 71.30 to 82.38 with o3-mini. The broader evaluation on five additional benchmarks shows a similar pattern: with Qwen-3.5-9B, the best CLEAR configurations outperform the strongest conventional baselines on MedBullets, MedExQA, AfriMedQA, MMLU, and MMLU-Pro, with gains ranging from 2.94 to 11.36 percentage points.

The two free-text benchmarks behave differently. On MedRBench, CLEAR improves over both Direct inference and the MedRAG baselines under both backbones, and the improvement holds on the Diagnosis and Treatment subsets alike; QPIS also performs well, slightly below the default CLEAR configuration. On HealthBench, the pattern depends on the backbone. Inspection of the intermediate outputs indicates that the o3-mini Direct baseline is already highly competitive on this benchmark and often produces a sufficiently strong answer without external grounding. In this setting, additional local or dynamic evidence contributes limited complementary information and can interfere with an otherwise correct response, which is consistent with both MedRAG and the default CLEAR configuration scoring below Direct inference. QPIS recovers to the level of Direct inference, suggesting that query planning and iterative search integrate external evidence more selectively. With qwen-3.5-9B the default configuration improves over Direct inference, from 0.3179 to 0.3444. Taken together, these results suggest that the benefit of cross-source adjudication depends on how far the parametric model already resolves the question on its own, and that retrieval does not necessarily help when it does.

The ablation results are consistent with a benefit from combining and adjudicating evidence across sources rather than simply relying on additional retrieved evidence. Across the o3-mini MCQ benchmarks, Direct inference, local BM25 retrieval, and dynamic evidence alone achieve aggregate scores of 87.48, 86.99, and 83.20, respectively, whereas CLEAR reaches 88.34 with the full cross-source framework. The targeted follow-up mechanism also provides additional benefit: removing the follow-up search step reduces aggregate performance from 78.13 to 77.22 with Qwen-3.5-9B. Notably, only 12 of 123 triggered follow-up searches ultimately lead to an accepted revision, suggesting that additional retrieval is used selectively rather than automatically. These results collectively provide preliminary evidence that preserving source provenance, explicitly identifying disagreement, and selectively resolving cross-source conflicts can offer advantages over relying on a single knowledge source or simply introducing more retrieved information.

\subsection{Relation to existing literature}

CLEAR builds on a growing body of work on retrieval-augmented generation, adaptive retrieval, and agentic information seeking. RAG provides a general mechanism for supplementing parametric language models with externally retrieved knowledge~\cite{lewis2020retrieval}. In medicine, systematic evaluations have shown that retrieval quality, corpus selection, and retriever--generator alignment can substantially influence downstream performance~\cite{xiong2024benchmarking}. Medical RAG approaches have further introduced domain-specific retrieval and reasoning strategies. For example, the knowledge-graph-based MedRAG framework proposed by \cite{zhao2025medrag} incorporates knowledge-graph-elicited reasoning into healthcare question answering, while Self-BioRAG combines retrieval with self-reflection to improve biomedical reasoning~\cite{jeong2024improving}. Collectively, these studies have established external grounding as an important strategy for improving medical reasoning and have provided a strong foundation for studying how LLMs should interact with external knowledge.

A related line of work has substantially advanced the retrieval process itself by making it adaptive. Self-RAG learns when to retrieve and critique retrieved passages through self-reflection~\cite{asai2024self}; Adaptive-RAG adjusts retrieval and reasoning strategies according to question complexity~\cite{jeong2024adaptive}; FLARE triggers retrieval during generation when the model anticipates missing information~\cite{jiang2023active}; and corrective RAG evaluates retrieved documents and acquires additional information when the initial retrieval is insufficient~\cite{yan2024corrective}. Medical extensions similarly use iterative retrieval, query reformulation, and follow-up questions to improve evidence acquisition~\cite{xiong2024improving,sohn2025rationale}. These studies demonstrate that retrieval need not be a fixed operation and provide important mechanisms for deciding whether, when, and how additional information should be acquired. CLEAR builds on this insight and focuses on a complementary question that becomes important once multiple knowledge sources are simultaneously available: how should a system determine whether an existing conclusion should be preserved, revised, or subjected to further evidence acquisition when those sources disagree?

Recent agentic systems further extend this paradigm by allowing models to autonomously invoke tools, rerank evidence, and iteratively acquire information. \cite{jia2026agentic} combines retrieval, reranking, evidence grounding, and memory mechanisms within an agentic medical question-answering workflow, while AgentClinic evaluates the ability of clinical agents to use retrieval, tools, and reflection in simulated clinical environments~\cite{schmidgall2026agentclinic}. These studies represent important steps toward more autonomous and adaptive medical reasoning systems and demonstrate the value of allowing agents to actively interact with external information. CLEAR is complementary to these efforts. Rather than focusing primarily on the orchestration of tools and retrieval actions, it explicitly maintains parametric, local, and dynamic evidence as source-specific candidate pathways before aggregation. Preserving this provenance makes agreement and disagreement among heterogeneous sources directly observable and allows these relationships to inform subsequent adjudication.

The proposed framework is most closely related to recent work on knowledge conflict, which has directly highlighted the limitations of treating either internal or external knowledge as inherently reliable. HealthContradict demonstrates that biomedical LLM behavior can change substantially when supplied evidence is correct, incorrect, or internally contradictory~\cite{zhang2026healthcontradict}, providing an important benchmark for understanding model behavior under conflicting evidence. Astute RAG further shows that neither internal nor external knowledge is consistently correct under conflict and develops mechanisms for managing imperfect retrieval and knowledge disagreement~\cite{wang2025astute}. \cite{wu2026conflict} advance this direction by using disagreement among candidate responses to guide additional retrieval across successive rounds and move the system toward better-supported consensus. CLEAR builds on this emerging line of research while considering a somewhat different evidence setting. Its candidate pathways are heterogeneous by construction: one reflects parametric knowledge, one is conditioned on a locally curated corpus, and one is grounded in dynamically acquired evidence. The framework therefore treats disagreement not only as a signal that additional retrieval may be needed, but also as information about how different evidence sources support competing conclusions. CLEAR first evaluates whether the available evidence is sufficient to preserve or revise the current conclusion, uses the override guard and challenge audit to examine both directions of potential error, and reserves additional search for conflicts that remain unresolved.

\subsection{Limitations and future work}

This study has several limitations. First, although we systematically evaluated CLEAR on ten benchmarks spanning both multiple-choice and free-text settings, the limitations of benchmark-based evaluation are well documented~\cite{wang2026beyond,chen2025benchmarking,bedi2025testing}. In particular, public benchmarks may overlap with data encountered during model pretraining or post-training~\cite{li2026memorization,alber2025medical}. We carefully audited the retrieved evidence for direct benchmark-item leakage in this study; however, the training data of the evaluated backbone models are not fully available, and potential parametric memorization therefore cannot be completely ruled out. This remains an open challenge for the evaluation of contemporary LLMs. Future work should prioritize validation on newly collected datasets to further reduce the risk of training-data overlap and associated memorization effects.

Second, CLEAR is designed as a general, training-free framework that can be applied to different backbone models without task-specific fine-tuning. Our experiments with both a commercial and an open-weight backbone provide preliminary evidence for this plug-and-play design, but the current adjudication mechanisms rely primarily on prompting and structured inference-time decision rules. The approach could potentially be further improved through post-training strategies, such as preference optimization or learning from expert adjudication examples. Evaluating such learned adjudication mechanisms while preserving generalizability across models and tasks represents an important direction for future work.

Third, our experiments remain offline evaluations and do not establish clinical effectiveness, safety, or readiness for deployment. Dynamic evidence acquisition introduces additional challenges in real clinical environments, including variability in search results, provenance and reliability of external sources, reproducibility over time, institutional access restrictions, privacy, and data-governance requirements~\cite{rahul2026regulatory}. These considerations may substantially affect how dynamic or other external evidence can be integrated into clinical workflows. Future studies should evaluate the framework in clinically relevant settings, including institutionally governed evidence environments.

Finally, although we evaluated different backbone models, retrieval configurations, and dynamic search implementations, broader evaluation across model families, retrievers, search systems, and application domains will be necessary to establish the generalizability of the findings. We make the implementation code publicly available to facilitate independent evaluation, extension, and refinement of the framework by the research community.
\section*{Data availability}
All datasets used in this study are publicly available. The multiple-choice QA benchmarks include MedQA, PubMedQA, NEJM-QA, MedBullets, MedExQA, AfriMedQA, MMLU, and MMLU-Pro. The free-text QA benchmarks include MedRBench and HealthBench. For retrieval-based experiments, we use the publicly available MedCorp corpus as the local static knowledge base, following the same corpus setting used in the MedRAG baseline. Online evidence is acquired at inference time through the web search interface described in the Methods section. No private clinical records, protected health information, or non-public patient-level data are used in this study.

\section*{Code availability}
The code used to implement the proposed framework, including multi-path candidate generation, evidence aggregation, online evidence acquisition, and evaluation scripts, will be available at https://github.com/Yale-BIDS-Chen-Lab/Agentic-Adjudication-for-Medical-Reasoning. All benchmarks are publicly available.

\section*{Acknowledgements}

This study is supported by the National Institutes of Health National Library of Medicine under Award Number R01LM014604.

\bibliographystyle{plainnat}
\bibliography{references}

@article{tian2024opportunities,
  title={Opportunities and challenges for ChatGPT and large language models in biomedicine and health},
  author={Tian, Shubo and Jin, Qiao and Yeganova, Lana and Lai, Po-Ting and Zhu, Qingqing and Chen, Xiuying and Yang, Yifan and Chen, Qingyu and Kim, Won and Comeau, Donald C and others},
  journal={Briefings in Bioinformatics},
  volume={25},
  number={1},
  pages={bbad493},
  year={2024},
  publisher={Oxford University Press}
}

@article{shi2024enhancing,
  title={Enhancing retrieval and managing retrieval: A four-module synergy for improved quality and efficiency in rag systems},
  author={Shi, Yunxiao and Zi, Xing and Shi, Zijing and Zhang, Haimin and Wu, Qiang and Xu, Min},
  journal={arXiv preprint arXiv:2407.10670},
  year={2024}
}

@inproceedings{zhang2023large,
  title={How do large language models capture the ever-changing world knowledge? a review of recent advances},
  author={Zhang, Zihan and Fang, Meng and Chen, Ling and Namazi-Rad, Mohammad-Reza and Wang, Jun},
  booktitle={Proceedings of the 2023 conference on empirical methods in natural language processing},
  pages={8289--8311},
  year={2023}
}

@inproceedings{wang2025bring,
  title={Bring your own knowledge: a survey of methods for LLM knowledge expansion},
  author={Wang, Mingyang and Stoll, Alisa and Lange, Lukas and Adel, Heike and Sch{\"u}tze, Hinrich and Str{\"o}tgen, Jannik},
  booktitle={Proceedings of the First Workshop on Large Language Model Memorization (L2M2)},
  pages={150--168},
  year={2025}
}

@article{tan2025dynamic,
  title={Dynamic parametric retrieval augmented generation for test-time knowledge enhancement},
  author={Tan, Yuqiao and He, Shizhu and Liao, Huanxuan and Zhao, Jun and Liu, Kang},
  journal={arXiv preprint arXiv:2503.23895},
  year={2025}
}

@article{wong2025retrieval,
  title={Retrieval-augmented systems can be dangerous medical communicators},
  author={Wong, Lionel and Ali, Ayman and Xiong, Raymond and Shen, Shannon Zeijang and Kim, Yoon and Agrawal, Monica},
  journal={arXiv preprint arXiv:2502.14898},
  year={2025}
}

@article{yang2026retrieval,
  title={Retrieval-augmented generation in medicine: A scoping review of technical implementations, clinical applications, and ethical considerations},
  author={Yang, Rui and Wong, Matthew Yu Heng and Li, Huitao and Li, Xin and Zhu, Wentao and Liao, Jingchi and Yu, Kunyu and Liew, Jonathan Chong Kai and Xuan, Weihao and Chen, Yingjian and others},
  journal={Cell Reports Medicine},
  year={2026},
  publisher={Elsevier}
}

@article{liu2025improving,
  title={Improving large language model applications in biomedicine with retrieval-augmented generation: a systematic review, meta-analysis, and clinical development guidelines},
  author={Liu, Siru and McCoy, Allison B and Wright, Adam},
  journal={Journal of the American Medical Informatics Association},
  volume={32},
  number={4},
  pages={605--615},
  year={2025},
  publisher={Oxford University Press}
}

@article{amugongo2025retrieval,
  title={Retrieval augmented generation for large language models in healthcare: A systematic review},
  author={Amugongo, Lameck Mbangula and Mascheroni, Pietro and Brooks, Steven and Doering, Stefan and Seidel, Jan},
  journal={PLOS Digital Health},
  volume={4},
  number={6},
  pages={e0000877},
  year={2025},
  publisher={Public Library of Science San Francisco, CA USA}
}

@article{guan2026large,
  title={Large Language Models Lack Temporal Awareness of Medical Knowledge},
  author={Guan, Zihan and Jin, Qiao and Xiong, Guangzhi and Chen, Fangyuan and Hu, Mengxuan and Chen, Qingyu and Peng, Yifan and Lu, Zhiyong and Vullikanti, Anil},
  journal={arXiv preprint arXiv:2605.13045},
  year={2026}
}

@article{artsi2025challenges,
  title={Challenges of implementing llms in clinical practice: Perspectives},
  author={Artsi, Yaara and Sorin, Vera and Glicksberg, Benjamin S and Korfiatis, Panagiotis and Freeman, Robert and Nadkarni, Girish N and Klang, Eyal},
  journal={Journal of Clinical Medicine},
  volume={14},
  number={17},
  pages={6169},
  year={2025},
  publisher={MDPI}
}

@article{kim2026medpmc,
  title={MedPMC: A Systematic Framework for Scaling High-Fidelity Medical Multimodal Data for Foundation Models},
  author={Kim, Hyunjae and Kim, Dain and Xiao, Pan and Applebaum, Serina S and Chung, Younjoon and Ai, Xuguang and Yin, Yu and Jiang, Roy and Du, Yuexi and Wei, Yawen and others},
  journal={arXiv preprint arXiv:2607.07673},
  year={2026}
}

@article{rahul2026regulatory,
  title={Regulatory Compliance and Data Governance in AI-Driven Healthcare: Legal and Regulatory Considerations for AI-Driven Healthcare Solutions},
  author={Rahul, SS and Kakade, Satish V},
  journal={Artificial Intelligence and Machine Learning in Neurology},
  volume={1},
  pages={79--108},
  year={2026},
  publisher={Wiley Online Library}
}

@article{wang2026reasoning,
  title={Reasoning-driven large language models in medicine: opportunities, challenges, and the road ahead},
  author={Wang, Xiaofei and Xiong, Zhuxin and Zou, Ke and Srinivasan, Sahana and Lo, Thaddaeus Wai Soon and Wu, Yilan and Zou, Minjie and Liu, Nan and Antaki, Fares and Ma, Weizhi and others},
  journal={The Lancet Digital Health},
  year={2026},
  publisher={Elsevier}
}

@article{alber2025medical,
  title={Medical large language models are vulnerable to data-poisoning attacks},
  author={Alber, Daniel Alexander and Yang, Zihao and Alyakin, Anton and Yang, Eunice and Rai, Sumedha and Valliani, Aly A and Zhang, Jeff and Rosenbaum, Gabriel R and Amend-Thomas, Ashley K and Kurland, David B and others},
  journal={Nature Medicine},
  volume={31},
  number={2},
  pages={618--626},
  year={2025},
  publisher={Nature Publishing Group US New York}
}

@article{li2026memorization,
  title={Memorization in large language models in medicine prevalence characteristics and implications},
  author={Li, Anran and Qian, Lingfei and Du, Mengmeng and Yin, Yu and Hu, Yan and Sun, Zihao and Fu, Yihang and Kim, Hyunjae and Stutz, Erica and Ai, Xuguang and others},
  journal={Nature Communications},
  year={2026},
  publisher={Nature Publishing Group}
}

@article{bedi2025testing,
  title={Testing and evaluation of health care applications of large language models: a systematic review},
  author={Bedi, Suhana and Liu, Yutong and Orr-Ewing, Lucy and Dash, Dev and Koyejo, Sanmi and Callahan, Alison and Fries, Jason A and Wornow, Michael and Swaminathan, Akshay and Lehmann, Lisa Soleymani and others},
  journal={Jama},
  volume={333},
  number={4},
  pages={319--328},
  year={2025}
}

@inproceedings{wang2026beyond,
  title={Beyond the leaderboard: Rethinking medical benchmarks for large language models},
  author={Wang, Wenxuan and Ma, Zizhan and Yu, Guo and Cheung, Yiu-Fai and Ding, Meidan and Liu, Jie and Chen, Wenting and Shen, Linlin},
  booktitle={Proceedings of the 64th Annual Meeting of the Association for Computational Linguistics (Volume 1: Long Papers)},
  pages={43078--43123},
  year={2026}
}

@article{chen2025benchmarking,
  title={Benchmarking large language models for biomedical natural language processing applications and recommendations},
  author={Chen, Qingyu and Hu, Yan and Peng, Xueqing and Xie, Qianqian and Jin, Qiao and Gilson, Aidan and Singer, Maxwell B and Ai, Xuguang and Lai, Po-Ting and Wang, Zhizheng and others},
  journal={Nature communications},
  volume={16},
  number={1},
  pages={3280},
  year={2025},
  publisher={Nature Publishing Group UK London}
}

@article{xie2025medical,
  title={Medical foundation large language models for comprehensive text analysis and beyond},
  author={Xie, Qianqian and Chen, Qingyu and Chen, Aokun and Peng, Cheng and Hu, Yan and Lin, Fongci and Peng, Xueqing and Huang, Jimin and Zhang, Jeffrey and Keloth, Vipina and others},
  journal={NPJ digital medicine},
  volume={8},
  number={1},
  pages={141},
  year={2025},
  publisher={Nature Publishing Group UK London}
}

@article{chen2026llm,
  title={LLM-assisted systematic review of large language models in clinical medicine},
  author={Chen, Sully F and Alyakin, Anton and Seas, Andreas and Yang, Eunice and Choi, Joanne J and Lee, Jin Vivian and Chen, Amelia L and Warman, Pranav I and Bitolas, Rochelle T and Steele, Robert J and others},
  journal={Nature medicine},
  volume={32},
  number={3},
  pages={1152},
  year={2026}
}

@article{liu2025application,
  title={Application of large language models in medicine},
  author={Liu, Fenglin and Zhou, Hongjian and Gu, Boyang and Zou, Xinyu and Huang, Jinfa and Wu, Jinge and Li, Yiru and Chen, Sam S and Hua, Yining and Zhou, Peilin and others},
  journal={Nature Reviews Bioengineering},
  volume={3},
  number={6},
  pages={445--464},
  year={2025},
  publisher={Nature Publishing Group UK London}
}

@article{singhal2023large,
  title={Large language models encode clinical knowledge},
  author={Singhal, Karan and Azizi, Shekoofeh and Tu, Tao and Mahdavi, S Sara and Wei, Jason and Chung, Hyung Won and Scales, Nathan and Tanwani, Ajay and Cole-Lewis, Heather and Pfohl, Stephen and others},
  journal={Nature},
  volume={620},
  number={7972},
  pages={172--180},
  year={2023},
  publisher={Nature Publishing Group UK London}
}

@article{chen2024meditron,
  title={Meditron-70b: Scaling medical pretraining for large language models},
  author={Chen, Zeming and Cano, Alejandro Hern{\'a}ndez and Romanou, Angelika and Bonnet, Antoine and Matoba, Kyle and Salvi, Francesco and Pagliardini, Matteo and Fan, Simin and K{\"o}pf, Andreas and Mohtashami, Amirkeivan and others},
  journal={arXiv preprint arXiv:2311.16079},
  year={2023}
}

@article{bolton2024biomedlm,
  title={Biomedlm: A 2.7 b parameter language model trained on biomedical text},
  author={Bolton, Elliot and Venigalla, Abhinav and Yasunaga, Michihiro and Hall, David and Xiong, Betty and Lee, Tony and Daneshjou, Roxana and Frankle, Jonathan and Liang, Percy and Carbin, Michael and others},
  journal={arXiv preprint arXiv:2403.18421},
  year={2024}
}

@inproceedings{zhao2025medrag,
  title={Medrag: Enhancing retrieval-augmented generation with knowledge graph-elicited reasoning for healthcare copilot},
  author={Zhao, Xuejiao and Liu, Siyan and Yang, Su-Yin and Miao, Chunyan},
  booktitle={Proceedings of the ACM on Web Conference 2025},
  pages={4442--4457},
  year={2025}
}

@article{jeong2024improving,
  title={Improving medical reasoning through retrieval and self-reflection with retrieval-augmented large language models},
  author={Jeong, Minbyul and Sohn, Jiwoong and Sung, Mujeen and Kang, Jaewoo},
  journal={Bioinformatics},
  volume={40},
  number={Supplement\_1},
  pages={i119--i129},
  year={2024},
  publisher={Oxford University Press}
}

@inproceedings{asai2024self,
  title={Self-rag: Learning to retrieve, generate, and critique through self-reflection},
  author={Asai, Akari and Wu, Zeqiu and Wang, Yizhong and Sil, Avi and Hajishirzi, Hannaneh},
  booktitle={International conference on learning representations},
  volume={2024},
  pages={9112--9141},
  year={2024}
}

@inproceedings{jeong2024adaptive,
  title={Adaptive-rag: Learning to adapt retrieval-augmented large language models through question complexity},
  author={Jeong, Soyeong and Baek, Jinheon and Cho, Sukmin and Hwang, Sung Ju and Park, Jong C},
  booktitle={Proceedings of the 2024 Conference of the North American Chapter of the Association for Computational Linguistics: Human Language Technologies (Volume 1: Long Papers)},
  pages={7036--7050},
  year={2024}
}

@article{jin2021disease,
  title={What disease does this patient have? a large-scale open domain question answering dataset from medical exams},
  author={Jin, Di and Pan, Eileen and Oufattole, Nassim and Weng, Wei-Hung and Fang, Hanyi and Szolovits, Peter},
  journal={Applied Sciences},
  volume={11},
  number={14},
  pages={6421},
  year={2021},
  publisher={MDPI}
}

@inproceedings{jin2019pubmedqa,
  title={Pubmedqa: A dataset for biomedical research question answering},
  author={Jin, Qiao and Dhingra, Bhuwan and Liu, Zhengping and Cohen, William and Lu, Xinghua},
  booktitle={Proceedings of the 2019 conference on empirical methods in natural language processing and the 9th international joint conference on natural language processing (EMNLP-IJCNLP)},
  pages={2567--2577},
  year={2019}
}

@article{wang2025baichuan,
  title={Baichuan-m1: Pushing the medical capability of large language models},
  author={Wang, Bingning and Zhao, Haizhou and Zhou, Huozhi and Song, Liang and Xu, Mingyu and Cheng, Wei and Zeng, Xiangrong and Zhang, Yupeng and Huo, Yuqi and Wang, Zecheng and others},
  journal={arXiv preprint arXiv:2502.12671},
  year={2025}
}

@article{qiu2025quantifying,
  title={Quantifying the reasoning abilities of LLMs on clinical cases},
  author={Qiu, Pengcheng and Wu, Chaoyi and Liu, Shuyu and Fan, Yanjie and Zhao, Weike and Chen, Zhuoxia and Gu, Hongfei and Peng, Chuanjin and Zhang, Ya and Wang, Yanfeng and others},
  journal={Nature Communications},
  volume={16},
  number={1},
  pages={9799},
  year={2025},
  publisher={Nature Publishing Group UK London}
}

@article{arora2025healthbench,
  title={Healthbench: Evaluating large language models towards improved human health},
  author={Arora, Rahul K and Wei, Jason and Hicks, Rebecca Soskin and Bowman, Preston and Qui{\~n}onero-Candela, Joaquin and Tsimpourlas, Foivos and Sharman, Michael and Shah, Meghan and Vallone, Andrea and Beutel, Alex and others},
  journal={arXiv preprint arXiv:2505.08775},
  year={2025}
}

@article{lewis2020retrieval,
  title={Retrieval-augmented generation for knowledge-intensive nlp tasks},
  author={Lewis, Patrick and Perez, Ethan and Piktus, Aleksandra and Petroni, Fabio and Karpukhin, Vladimir and Goyal, Naman and K{\"u}ttler, Heinrich and Lewis, Mike and Yih, Wen-tau and Rockt{\"a}schel, Tim and others},
  journal={Advances in neural information processing systems},
  volume={33},
  pages={9459--9474},
  year={2020}
}

@inproceedings{xiong2024benchmarking,
  title={Benchmarking retrieval-augmented generation for medicine},
  author={Xiong, Guangzhi and Jin, Qiao and Lu, Zhiyong and Zhang, Aidong},
  booktitle={Findings of the Association for Computational Linguistics: ACL 2024},
  pages={6233--6251},
  year={2024}
}

@misc{yan2024corrective,
      title={Corrective Retrieval Augmented Generation}, 
      author={Shi-Qi Yan and Jia-Chen Gu and Yun Zhu and Zhen-Hua Ling},
      year={2024},
      eprint={2401.15884},
      archivePrefix={arXiv},
      primaryClass={cs.CL},
      url={https://arxiv.org/abs/2401.15884}, 
}

@inproceedings{jiang2023active,
  title={Active retrieval augmented generation},
  author={Jiang, Zhengbao and Xu, Frank F and Gao, Luyu and Sun, Zhiqing and Liu, Qian and Dwivedi-Yu, Jane and Yang, Yiming and Callan, Jamie and Neubig, Graham},
  booktitle={Proceedings of the 2023 conference on empirical methods in natural language processing},
  pages={7969--7992},
  year={2023}
}

@inproceedings{xie2025retrieval,
  title={When Retrieval Hurts: A Critical Analysis of RAG in Medical Question Answering},
  author={Xie, Weidong and Fang, Yushan and Zheng, Xinxin and Liu, Dongjun and Li, Zihan},
  booktitle={Proceedings of the 2025 3rd International Conference on Artificial Intelligence, Systems and Network Security},
  pages={312--317},
  year={2025}
}

@inproceedings{xiong2024improving,
  title={Improving retrieval-augmented generation in medicine with iterative follow-up questions},
  author={Xiong, Guangzhi and Jin, Qiao and Wang, Xiao and Zhang, Minjia and Lu, Zhiyong and Zhang, Aidong},
  booktitle={Biocomputing 2025: Proceedings of the Pacific Symposium},
  pages={199--214},
  year={2024},
  organization={World Scientific}
}

@article{brown2020language,
  title={Language models are few-shot learners},
  author={Brown, Tom and Mann, Benjamin and Ryder, Nick and Subbiah, Melanie and Kaplan, Jared D and Dhariwal, Prafulla and Neelakantan, Arvind and Shyam, Pranav and Sastry, Girish and Askell, Amanda and others},
  journal={Advances in neural information processing systems},
  volume={33},
  pages={1877--1901},
  year={2020}
}

@article{chowdhery2023palm,
  title={Palm: Scaling language modeling with pathways},
  author={Chowdhery, Aakanksha and Narang, Sharan and Devlin, Jacob and Bosma, Maarten and Mishra, Gaurav and Roberts, Adam and Barham, Paul and Chung, Hyung Won and Sutton, Charles and Gehrmann, Sebastian and others},
  journal={Journal of machine learning research},
  volume={24},
  number={240},
  pages={1--113},
  year={2023}
}

@misc{llama2023llama,
      title={LLaMA: Open and Efficient Foundation Language Models}, 
      author={Hugo Touvron and Thibaut Lavril and Gautier Izacard and Xavier Martinet and Marie-Anne Lachaux and Timothée Lacroix and Baptiste Rozière and Naman Goyal and Eric Hambro and Faisal Azhar and Aurelien Rodriguez and Armand Joulin and Edouard Grave and Guillaume Lample},
      year={2023},
      eprint={2302.13971},
      archivePrefix={arXiv},
      primaryClass={cs.CL},
      url={https://arxiv.org/abs/2302.13971}, 
}

@article{achiam2023gpt,
  title={Gpt-4 technical report},
  author={Achiam, Josh and Adler, Steven and Agarwal, Sandhini and Ahmad, Lama and Akkaya, Ilge and Aleman, Florencia Leoni and Almeida, Diogo and Altenschmidt, Janko and Altman, Sam and Anadkat, Shyamal and others},
  journal={arXiv preprint arXiv:2303.08774},
  year={2023}
}

@article{singhal2025toward,
  title={Toward expert-level medical question answering with large language models},
  author={Singhal, Karan and Tu, Tao and Gottweis, Juraj and Sayres, Rory and Wulczyn, Ellery and Amin, Mohamed and Hou, Le and Clark, Kevin and Pfohl, Stephen R and Cole-Lewis, Heather and others},
  journal={Nature medicine},
  volume={31},
  number={3},
  pages={943--950},
  year={2025},
  publisher={Nature Publishing Group US New York}
}

@article{dhingra2022time,
  title={Time-aware language models as temporal knowledge bases},
  author={Dhingra, Bhuwan and Cole, Jeremy R and Eisenschlos, Julian Martin and Gillick, Daniel and Eisenstein, Jacob and Cohen, William W},
  journal={Transactions of the Association for Computational Linguistics},
  volume={10},
  pages={257--273},
  year={2022},
  publisher={MIT Press One Broadway, 12th Floor, Cambridge, Massachusetts 02142, USA~…}
}

@misc{nlm2024medline,
  author       = {{National Library of Medicine}},
  title        = {MEDLINE PubMed Production Statistics},
  year         = {2024},
  url          = {https://www.nlm.nih.gov/bsd/medline_pubmed_production_stats.html},
  note         = {Last reviewed April 30, 2024}
}

@article{zakka2024almanac,
  title={Almanac—retrieval-augmented language models for clinical medicine},
  author={Zakka, Cyril and Shad, Rohan and Chaurasia, Akash and Dalal, Alex R and Kim, Jennifer L and Moor, Michael and Fong, Robyn and Phillips, Curran and Alexander, Kevin and Ashley, Euan and others},
  journal={Nejm ai},
  volume={1},
  number={2},
  pages={AIoa2300068},
  year={2024},
  publisher={Massachusetts Medical Society}
}

@inproceedings{sohn2025rationale,
  title={Rationale-guided retrieval augmented generation for medical question answering},
  author={Sohn, Jiwoong and Park, Yein and Yoon, Chanwoong and Park, Sihyeon and Hwang, Hyeon and Sung, Mujeen and Kim, Hyunjae and Kang, Jaewoo},
  booktitle={Proceedings of the 2025 Conference of the Nations of the Americas Chapter of the Association for Computational Linguistics: Human Language Technologies (Volume 1: Long Papers)},
  pages={12739--12753},
  year={2025}
}

@article{kim2025rethinking,
  title={Rethinking retrieval-augmented generation for medicine: A large-scale, systematic expert evaluation and practical insights},
  author={Kim, Hyunjae and Sohn, Jiwoong and Gilson, Aidan and Cochran-Caggiano, Nicholas and Applebaum, Serina and Jin, Heeju and Park, Seihee and Park, Yujin and Park, Jiyeong and Choi, Seoyoung and others},
  journal={arXiv preprint arXiv:2511.06738},
  year={2025}
}

@inproceedings{wang2025astute,
  title={Astute rag: Overcoming imperfect retrieval augmentation and knowledge conflicts for large language models},
  author={Wang, Fei and Wan, Xingchen and Sun, Ruoxi and Chen, Jiefeng and Arik, Sercan O},
  booktitle={Proceedings of the 63rd Annual Meeting of the Association for Computational Linguistics (Volume 1: Long Papers)},
  pages={30553--30571},
  year={2025}
}

@online{guidelinecentral2025review,
  author  = {{Guideline Central}},
  title   = {Clinical Practice Guidelines Published in 2024:
             Guidelines Year in Review},
  year    = {2025},
  date    = {2025-01-20},
  url     = {https://www.guidelinecentral.com/insights/2024-in-review-guidelines/},
  urldate = {2026-07-14}
}

@article{jia2026agentic,
  title={Agentic memory-augmented retrieval and evidence grounding for medical question-answering tasks},
  author={Jia, Shuyue and Bit, Subhrangshu and Jasodanand, Varuna H and Liu, Yi and Kolachalama, Vijaya B},
  journal={International Journal of Medical Informatics},
  pages={106339},
  year={2026},
  publisher={Elsevier}
}

@article{zhang2026healthcontradict,
  title={Healthcontradict: Evaluating biomedical knowledge conflicts in language models. npj Digital Medicine},
  author={Zhang, Boya and Bornet, Alban and Yang, Rui and Liu, Nan and Teodoro, Douglas},
  year={2026}
}

@article{schmidgall2026agentclinic,
  title={AgentClinic: a multimodal benchmark for tool-using clinical AI agents},
  author={Schmidgall, Samuel and Ziaei, Rojin and Harris, Carl and Kim, Ji Woong and Reis, Eduardo Pontes and Jopling, Jeffrey and Moor, Michael},
  journal={npj Digital Medicine},
  year={2026},
  publisher={Nature Publishing Group UK London}
}

@article{wu2026conflict,
  title={From conflict to consensus: Boosting medical reasoning via multi-round agentic rag},
  author={Wu, Wenhao and Tang, Zhentao and Li, Yafu and Kai, Shixiong and Yuan, Mingxuan and Sun, Zhenhong and Chen, Chunlin and Wang, Zhi},
  journal={arXiv preprint arXiv:2603.03292},
  year={2026}
}

@inproceedings{kim2024medexqa,
  title={MedExQA: Medical question answering benchmark with multiple explanations},
  author={Kim, Yunsoo and Wu, Jinge and Abdulle, Yusuf and Wu, Honghan},
  booktitle={Proceedings of the 23rd Workshop on biomedical natural language processing},
  pages={167--181},
  year={2024}
}

@inproceedings{nimo2025afrimed,
  title={AfriMed-QA: A Pan-African, multi-specialty, medical question-answering benchmark dataset},
  author={Nimo, Charles and Olatunji, Tobi and Owodunni, Abraham Toluwase and Abdullahi, Tassallah and Ayodele, Emmanuel and Sanni, Mardhiyah and Aka, Ezinwanne C and Omofoye, Folafunmi and Yuehgoh, Foutse and Faniran, Timothy and others},
  booktitle={Proceedings of the 63rd Annual Meeting of the Association for Computational Linguistics (Volume 1: Long Papers)},
  pages={1948--1973},
  year={2025}
}

@article{hendrycks2020measuring,
  title={Measuring massive multitask language understanding},
  author={Hendrycks, Dan and Burns, Collin and Basart, Steven and Zou, Andy and Mazeika, Mantas and Song, Dawn and Steinhardt, Jacob},
  journal={arXiv preprint arXiv:2009.03300},
  year={2020}
}

@article{wang2024mmlu,
  title={Mmlu-pro: A more robust and challenging multi-task language understanding benchmark},
  author={Wang, Yubo and Ma, Xueguang and Zhang, Ge and Ni, Yuansheng and Chandra, Abhranil and Guo, Shiguang and Ren, Weiming and Arulraj, Aaran and He, Xuan and Jiang, Ziyan and others},
  journal={Advances in Neural Information Processing Systems},
  volume={37},
  pages={95266--95290},
  year={2024}
}

\appendix
\appendix

\section{Additional Experimental Details}
\label{app:details}

\subsection{Additional Results}
\label{sec:app_additional_results}

To further examine the stability of CLEAR beyond the datasets used in our main evaluation, we additionally evaluate the framework on five benchmarks spanning medical question answering and broader knowledge and reasoning tasks. MedBullets contains 308 USMLE Step 2 and Step 3-style multiple-choice questions with expert-written explanations, collected from the MedBullets platform~\cite{chen2025benchmarking}. MedExQA targets five medical specialties underrepresented in existing benchmarks and pairs each question with multiple reference explanations to support explanation-quality evaluation~\cite{kim2024medexqa}. AfriMed-QA is a large-scale, Pan-African, multi-specialty medical question-answering dataset spanning more than 60 medical schools across 16 countries~\cite{nimo2025afrimed}. MMLU is a broad multitask benchmark of four-option multiple-choice questions across 57 subjects spanning STEM, humanities, social sciences, and professional domains~\cite{hendrycks2020measuring}. MMLU-Pro extends MMLU with more reasoning-focused questions, ten answer choices instead of four, and broader coverage across 14 domains, including health~\cite{wang2024mmlu}.

\subsubsection{Results on Additional Datasets}




\begin{table*}[t]
\centering
\caption{Results on additional MCQ datasets.}
\label{tab:additional_mcq}
\resizebox{\textwidth}{!}{
\begin{tabular}{llccccc}
\toprule
\textbf{Method} & \textbf{Backbone}
& \textbf{MedBullets}
& \textbf{MedExQA}
& \textbf{AfriMedQA}
& \textbf{MMLU}
& \textbf{MMLU-Pro} \\
\midrule

\multicolumn{7}{l}{\textbf{GPT-based Model}} \\
\midrule
Direct
& gpt-o3-mini
& 81.49 & 85.45 & 79.89 & 92.65 & 77.02 \\

MedRAG (BM25 Setting)
& gpt-o3-mini
& 80.52 & 84.06 & 76.44 & 93.39 & 75.92 \\

MedRAG (MedCPT Setting)
& gpt-o3-mini
& 81.17 & 85.03 & 78.74 & 92.84 & 74.94 \\

CLEAR
& gpt-o3-mini
& 81.82 & \textbf{86.20} & 78.74 & 93.85 & 76.89 \\

CLEAR (dual models)
& gpt-o3-mini
& \textbf{83.77} & 86.10 & 79.89 & 93.76 & 77.26 \\

CLEAR (Query Planning + Iterative Search)
& gpt-o3-mini
& 83.12 & 85.99 & \textbf{79.89} & \textbf{93.85} & \textbf{78.00} \\

\midrule
\multicolumn{7}{l}{\textbf{Qwen-based Model}} \\
\midrule
Direct
& qwen-3.5-9B
& 52.60 & 79.57 & 70.69 & 85.22 & 65.65 \\

MedRAG (BM25 Setting)
& qwen-3.5-9B
& 53.90 & 79.25 & 70.11 & 86.96 & 66.38 \\

MedRAG (MedCPT Setting)
& qwen-3.5-9B
& 54.22 & 80.86 & 68.39 & 86.59 & 66.50 \\

CLEAR
& qwen-3.5-9B
& \textbf{65.58} & 83.53 & 75.29 & 89.72 & \textbf{72.00} \\

CLEAR (dual models)
& qwen-3.5-9B
& 63.96 & \textbf{83.96} & 76.44 & 89.72 & 71.52 \\

CLEAR (Query Planning + Iterative Search)
& qwen-3.5-9B
& 63.64 & 83.21 & \textbf{77.59} & \textbf{89.90} & 70.78 \\

\bottomrule
\end{tabular}
}
\end{table*}

\begin{figure}[htbp]
    \centering
    \includegraphics[width=\textwidth]{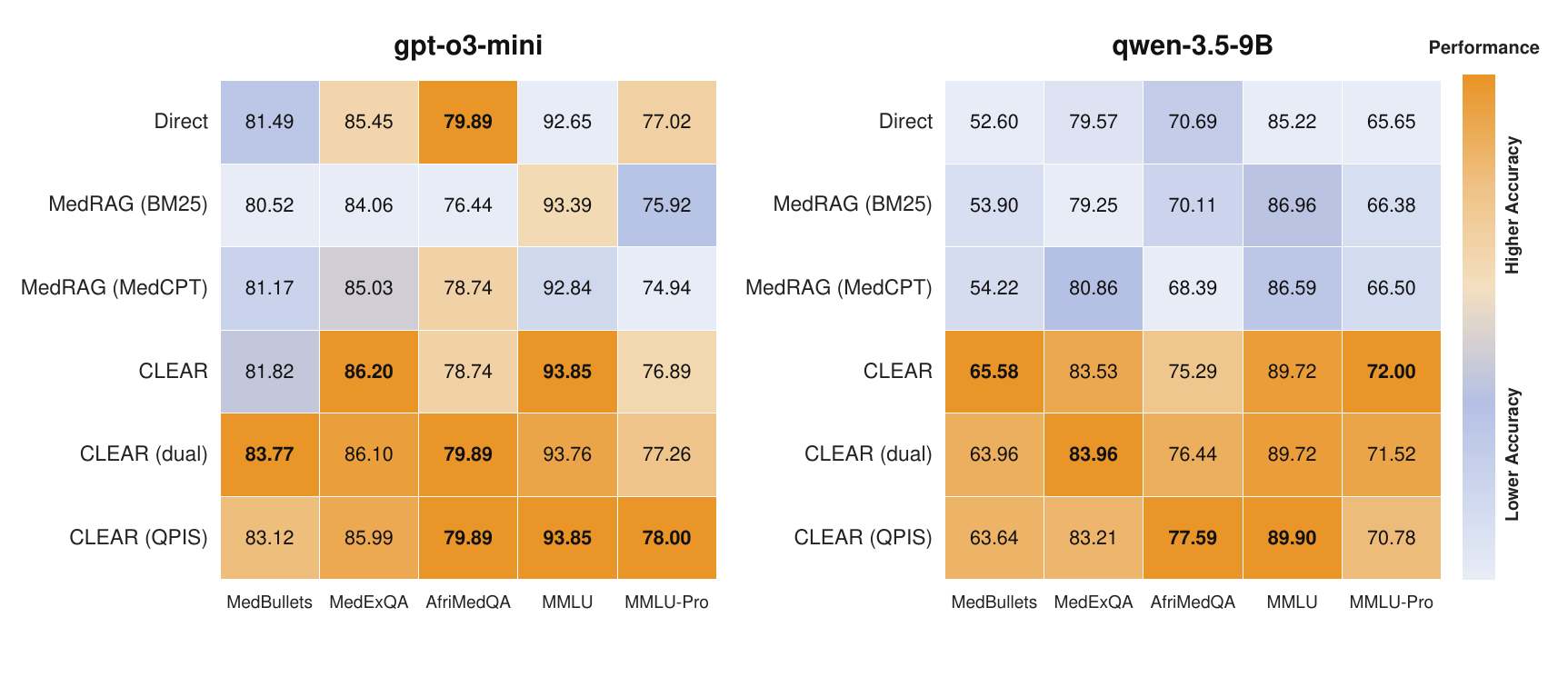}
    \caption{Overall results on five additional benchmarks. Performance is shown for each method and backbone, with the color scale normalized separately within each dataset column.}
    \label{fig:additional_mcq}
\end{figure}

As shown in Figure~\ref{fig:additional_mcq}, CLEAR remains competitive across the five additional benchmarks. With the Qwen-3.5-9B backbone, the best CLEAR configuration outperforms the strongest evaluated baseline on all five benchmarks, with gains ranging from 2.94 to 11.36 percentage points. With the o3-mini backbone, at least one CLEAR configuration matches or exceeds the strongest baseline on each benchmark. These results provide additional evidence that the framework can be applied across both medical question-answering tasks and broader knowledge and reasoning benchmarks.

\subsection{Implementation Details}

\paragraph{Backbone configuration.}
We instantiate CLEAR with two backbones: o3-mini as a commercial model and Qwen-3.5-9B as an open-weight model. o3-mini is accessed through the Azure OpenAI API (version 2024-12-01-preview), with temperature and reasoning effort left at their default settings and a request timeout of 120 seconds. Qwen-3.5-9B is served through an OpenAI-compatible server and queried with temperature 0, top-$p$ 0.95, and a maximum of 1024 output tokens, with a request timeout of 300 seconds. In the primary configuration, the same backbone $\mathcal{M}$ performs candidate generation across all three pathways, source-quality assessment, aggregation verification, override-guard and challenge-audit adjudication, and task-specific response generation.

\paragraph{Candidate generation.}
The parametric pathway queries $\mathcal{M}$ without retrieval. The local pathway uses MedCorp~\cite{xiong2024benchmarking} as $K_{\mathrm{local}}$, the same corpus used by the MedRAG baselines, so that local retrieval is compared using identical source material. MedCorp aggregates PubMed (23.9M documents), 18 medical textbooks, and Wikipedia (6.5M articles). We instantiate $\mathcal{R}$ as BM25 and retrieve the top $k=16$ snippets. The framework is agnostic to the choice of $\mathcal{R}$, and BM25 is used here as a controlled sparse-retrieval instantiation. The dynamic pathway acquires evidence through web search. The primary experiments use GPT-4o's web-search capability for dynamic evidence acquisition. As an additional implementation-diversity experiment, we instantiate the dynamic pathway using Tavily on the NEJM-QA Internal Medicine subset, providing preliminary evidence that CLEAR can also be implemented with an alternative web-search backend.

\paragraph{Source-quality assessment.}
The eight retrieved dynamic documents are assessed in a single backbone call and assigned a categorical quality level of high, medium, or low according to source authority, recency, and content type. The resulting counts, denoted $n_{\mathrm{high}}$, $n_{\mathrm{medium}}$, and $n_{\mathrm{low}}$, constitute the source-quality information $S_{\mathrm{dynamic}}$ used by the aggregation verifier and adjudication modules. We use categorical rather than continuous quality scores to avoid introducing an additional manually calibrated threshold.

\paragraph{Aggregation verifier.}
The verifier receives the query, the option set for multiple-choice tasks, the three candidate records, the local and dynamic evidence, and $S_{\mathrm{dynamic}}$, and returns a structured decision containing the preliminary answer, the routing decision $z_0 \in \mathcal{Z}$, a rationale, and a confidence signal.

\paragraph{Override guard.}
The override guard is invoked when $z_0=\texttt{accept\_dynamic}$ and the dynamic candidate conflicts with a competing conclusion. In particular, when
$a_{\mathrm{param}}=a_{\mathrm{local}}\neq a_{\mathrm{dynamic}}$,
the parametric--local consensus is treated as the default; when all three candidates differ, the parametric candidate is used as the default for the guard assessment.

A targeted follow-up search is issued only when the dynamic evidence includes at least one high-quality source ($n_{\mathrm{high}}\geq1$); otherwise, the proposed override is rejected without further retrieval. The override is granted only when the follow-up assessment supports the challenging conclusion, establishes that the default answer is not better supported, confirms that the retrieved evidence is aligned with the query, provides support at the exact option level when applicable, and leaves no material unresolved conflict. When these conditions are not satisfied, the system returns the default answer with route \texttt{keep\_local} or \texttt{keep\_parametric}, as appropriate.

\paragraph{Challenge audit.}
The challenge audit is complementary to the override guard and is invoked when the verifier does \emph{not} initially accept the dynamic candidate, but the dynamic pathway challenges an existing parametric--local consensus,
$a_{\mathrm{param}}=a_{\mathrm{local}}\neq a_{\mathrm{dynamic}}$.
In this case, the existing consensus is treated as the default and the dynamic answer as the challenger.

A targeted follow-up search is used to determine whether sufficient evidence exists to reopen the decision. The challenge is considered only when the retrieved evidence contains at least one high-quality source or at least two medium-quality sources
($n_{\mathrm{high}}\geq1$ or $n_{\mathrm{medium}}\geq2$).
If this evidence requirement is not met, or if the follow-up evidence does not clearly support the challenger, the original parametric--local consensus is preserved.

\paragraph{Task adapters.}
For multiple-choice tasks, the format gate converts the adjudicated conclusion into the option representation required by each benchmark and supports both single-answer and multiple-answer settings, the latter arising in NEJM-QA. For free-text tasks, the response generator composes the final answer from the adjudicated conclusion, its rationale, and the evidence bundle, without repeating retrieval or revising the adjudication result.

\paragraph{Implementation diversity.}

As a complementary implementation-diversity analysis, we instantiate the dynamic-evidence pathway using the Tavily API on the 126-question NEJM-QA Internal Medicine subset with Qwen-3.5-9B. This experiment is intended to assess whether CLEAR can be instantiated with an alternative web-search backend rather than to provide a comprehensive comparison among search systems. No backend-specific performance tuning is applied. The results are shown in Table~\ref{tab:nejm_internal_medicine_results}.

\begin{table}[t]
  \centering
  \begin{tabular}{lc}
  \hline
  Method & Accuracy \\
  \hline
  Direct & 84/126 (66.7\%) \\
  BM25 & 80/126 (63.5\%) \\
  Dynamic only & 85/126 (67.5\%) \\
  CLEAR & 89/126 (70.6\%) \\
  \hline
  \end{tabular}
  \caption{Results on the NEJM-QA Internal Medicine subset using Qwen-3.5-9B and Tavily for dynamic evidence acquisition. The experiment provides preliminary evidence that CLEAR can be instantiated with an alternative web-search backend.}
  \label{tab:nejm_internal_medicine_results}
\end{table}

\end{document}